\documentclass[conference]{IEEEtran}
\IEEEoverridecommandlockouts
\usepackage{amsmath,graphicx}

\usepackage{times}
\usepackage{soul}
\usepackage{url}
\usepackage[utf8]{inputenc}
\usepackage[small]{caption}
\usepackage{epsfig}
\usepackage{url,bm,amsthm}
\usepackage{graphicx}
\usepackage{mathrsfs,balance}
\usepackage[ruled,vlined]{algorithm2e}
\usepackage{graphics,epstopdf,color,enumerate,subfigure,balance}
\graphicspath{{./figures/}}

\usepackage{caption}
\usepackage{makecell,amsmath,amsfonts,amssymb}
\usepackage{hyperref}
\usepackage{courier}
\usepackage{todonotes}
\usepackage{diagbox}

\makeatletter
\def\BibTeX{{\rm B\kern-.05em{\sc i\kern-.025em b}\kern-.08em
    T\kern-.1667em\lower.7ex\hbox{E}\kern-.125emX}}
\begin{document}
\title{SEEC: Semantic Vector Federation across Edge Computing Environments}

\author{
\IEEEauthorblockN{Shalisha Witherspoon\IEEEauthorrefmark{1},
Dean Steuer\IEEEauthorrefmark{1}, Graham Bent\IEEEauthorrefmark{2} and
Nirmit Desai\IEEEauthorrefmark{1}}
\IEEEauthorblockA{\IEEEauthorrefmark{1}IBM Research,
Yorktown Heights, NY, USA\\
Email: Shalisha.Witherspoon@ibm.com,
Dean.Steuer@ibm.com,
nirmit.desai@us.ibm.com}
\IEEEauthorblockA{\IEEEauthorrefmark{2}IBM Research,
Hursley, Winchester, UK\\
Email: GBent@uk.ibm.com}
}

\maketitle
\thispagestyle{plain}
\pagestyle{plain}

\begin{abstract}
Semantic vector embedding techniques have proven useful in learning semantic representations of data across multiple domains. A key application enabled by such techniques is the ability to measure semantic similarity between given data samples and find data most similar to a given sample. State-of-the-art embedding approaches assume all data is available on a single site. However, in many business settings, data is distributed across multiple edge locations and cannot be aggregated due to a variety of constraints.  Hence, the applicability of state-of-the-art embedding approaches is limited to freely shared datasets, leaving out applications with sensitive or mission-critical data. This paper addresses this gap by proposing novel unsupervised algorithms called \emph{SEEC} for learning and applying semantic vector embedding in a variety of distributed settings. Specifically, for scenarios where multiple edge locations can engage in joint learning, we adapt the recently proposed federated learning techniques for semantic vector embedding. Where joint learning is not possible, we propose novel semantic vector translation algorithms to enable semantic query across multiple edge locations, each with its own semantic vector-space. Experimental results on natural language as well as graph datasets show that this may be a promising new direction.
\end{abstract}


\section{Introduction}\label{sec:intro}
Exponential growth of IoT devices and the need to analyze the vast amounts of data they generate closer to its origin has led to an emergence of the \emph{edge computing} paradigm \cite{satyanarayanan-edge-2017}.  The factors driving such a paradigm shift are fundamental: (a) costs involved in transporting large amounts of data to Cloud, (b) regulatory constraints in moving data across sites, and (c) latency in placing all data analytics in Cloud. Further, deployments of applications enabled by 5G network architecture rely on edge computing for meeting the low-latency requirements\cite{hu2015mobile}. 

A killer application of edge computing is the extraction of insights from the edge data by running machine learning computations at the edge, without needing to export the data to a central location such as the Cloud \cite{Zhou-edge-intelligence-2019}. However, most of the recent advances in machine learning have focused on performance improvements while assuming all data is aggregated in a central location with massive computational capacity. Recently proposed federated learning techniques have charted a new direction by enabling model training from data residing locally across many edge locations \cite{yang-fed-concepts-2019,mcmahan-fed-averaging-2016}. 

However, previous work on federated learning has not investigated machine learning tasks beyond classification and prediction. Specifically, representation learning and semantic vector embedding techniques have proven effective across a variety of machine learning tasks across multiple domains. For text data, sentence and paragraph embedding techniques such as doc2vec \cite{le-doc2vec-2014}, GloVe \cite{pennington2014glove}, and BERT \cite{BERT-devlin-2018} have led to highly accurate language models for a variety of NLP tasks. Similar results have been achieved in graph learning tasks \cite{Grover-node2vec-2016,wang2017knowledge} and image recognition tasks \cite{norouzi-zeroshot-2014,devise-frome-2013}.  Key reasons behind the effectiveness of semantic embedding techniques include their ability to numerically represent rich features in low-dimension vectors and their ability to preserve semantic similarity among such rich features. Further, little or no labeled data is needed in learning the semantic vector embedding models. Clearly, semantic vector embedding will remain a fundamental tool in addressing many machine learning problems in the future.

This paper addresses the challenge of representation learning when data cannot be in one location. Two new research problems are introduced that generalize federated learning. First, we introduce the problem of learning semantic vector embedding wherein each edge site with data participates in an iterative joint learning process. However, unlike the previous work on federated learning, the edge sites must agree on the vector-space encoding. Second, we address a different setting where the edge sites are unable to participate in an iterative joint learning process. Instead, each edge site maintains a semantic vector embedding model of its own. Such scenarios are quite common where edge sites may not have continuous connectivity and may join and leave dynamically. We refer to both of these problems collectively as \emph{SEEC} -- the challenge of federating semantic vectors across edge sites.

This paper presents novel algorithms to address the SEEC challenge. In the case of joint learning, prior to beginning the iterative distributed gradient descent, edge sites collaborate to compute an aggregate feature set so that the semantic vectors spaces across edge sites are aligned. In the case where joint learning is not possible, edge sites learn their own semantic vector embedding models from local data. As a result, the semantic vector-spaces across edge sites are not aligned and semantic similarity across edge sites is not preserved. To address this problem, we propose a novel approach for learning a mapping function between the vector-spaces such that vectors of one edge site can be mapped to semantically similar vectors on another edge site.

We evaluate the SEEC algorithms in the context of a real-world application scenario of performing semantic search for resources across multiple collaborative edge locations wherein each edge location locally stores data about its resources and does not trust other edge locations with raw data. The key criteria for evaluation is the quality of semantic search results produced by the SEEC algorithms relative to those produced by the state-of-the-art centralized algorithms. We evaluate two data modalities in the experiments: (1) multiple natural language datasets for evaluating federated doc2vec semantic vector embedding and (2) a graph dataset for evaluating federated node2vec semantic vector embedding. It is important to note that the focus of this paper is in addressing the federated learning challenge in semantic vector embedding, regardless of the specific embedding technique being applied. Thus, doc2vec and node2vec are mere instruments that enable the experimental evaluation. The results show that the SEEC algorithms produce semantic search results that are quite similar to those produced by the traditional centralized approaches. We also find that the natural language datasets are more amenable to federated embedding than the graph datasets, primarily due to the asymmetry involved in partitioning graph data across edge sites.

In summary, this paper makes the following contributions:
\begin{itemize}
    \item Introduce a novel direction for edge computing research in addressing the challenge of representation learning in federated fashion
    \item Formulate two research problems in learning semantic vector embedding targeting two different settings wherein joint learning may or may not be feasible
    \item Present novel algorithms for iterative learning and vector-space translation to address the research problems
    \item Evaluate the algorithms with experiments on multi-modal real-world datasets and widely used embedding algorithms of doc2vec and node2vec relative to the scenario of semantically querying edge resources
\end{itemize}

The rest of this paper is structured as follows. Section~\ref{sec:background} introduces a motivating application scenario followed by a brief background on federated learning, doc2vec, and node2vec. Section~\ref{sec:problem} formulates the technical problems of federated learning of semantic vector embedding via joint learning as well as vector-space translation. Section~\ref{sec:algo} then presents the novel SEEC algorithms and Section~\ref{sec:exp} presents experimental setup and evaluation results. Lastly, Section~\ref{sec:related} presents related works and Section~\ref{sec:conclusion} concludes the paper.

\section{Background}\label{sec:background}
Without the loss of generality, this section describes an application scenario of finding resources across edge sites to motivate the challenge of federated semantic vector embedding, followed by a brief primer on federated learning, doc2vec, and node2vec algorithms.

\subsection{Motivating Example}\label{sec:motivation}
\begin{figure}[htb!]
  \includegraphics[width=\columnwidth]{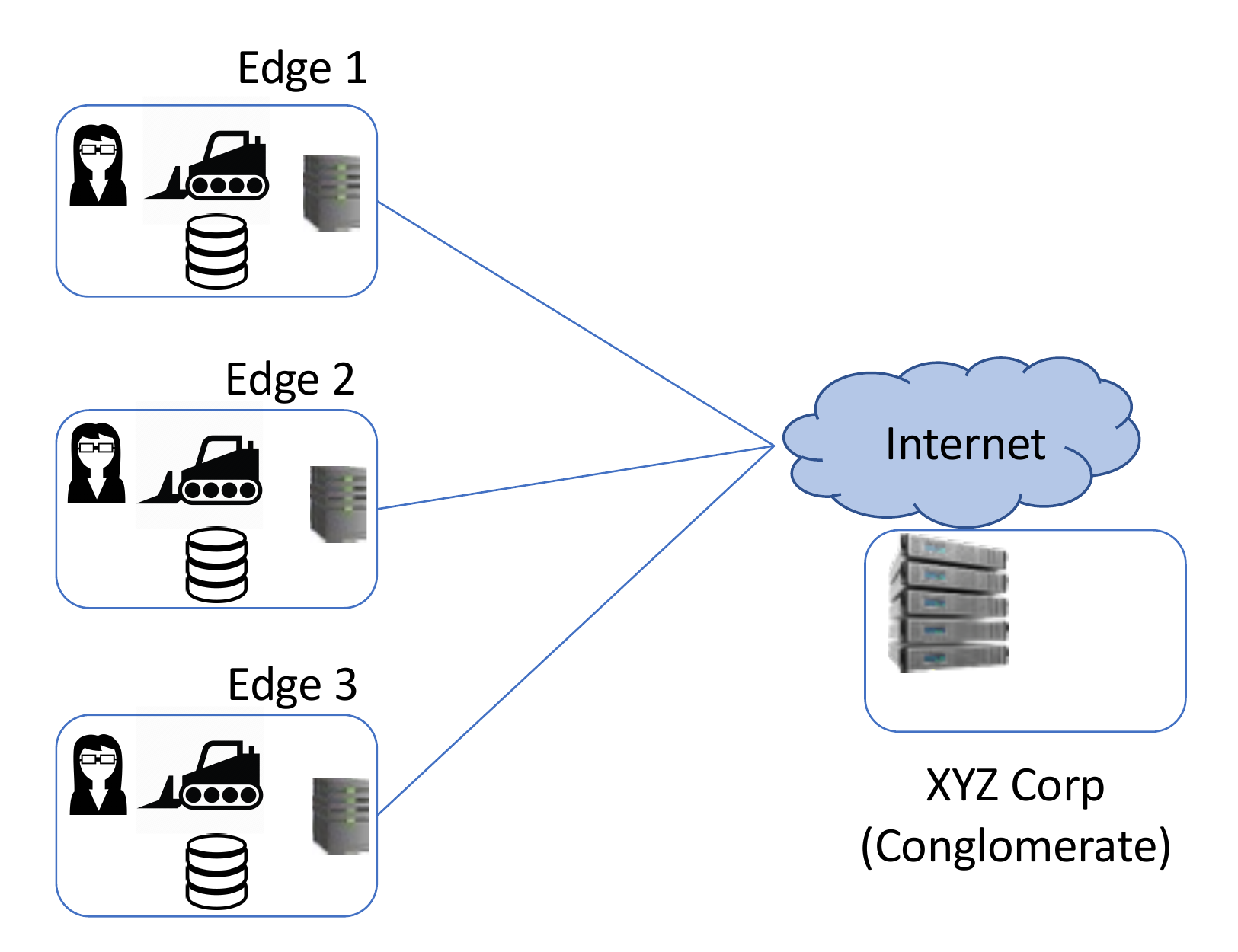}
  \caption{Motivating example: Search and share edge resources in a conglomerate}
  \label{fig:example}
\end{figure}
To help illustrate the challenges, we describe a motivating application scenario, without loss of general applicability of the techniques contributed by this paper. XYZ Corp is a business conglomerate with many subsidiary companies operating across multiple locations across the world.  The subsidiaries are autonomous, each focused on its domain of business. The conglomerate operates via a loosely coupled collaboration among the subsidiaries whose IT systems and data are siloed due to a variety of regulatory requirements, customization of such systems according to business needs, and the complexity of integrating them into a single system. Hence, each subsidiary maintains locally its own operational data produced by business process resources (people, equipment) and other information assets and is unable to share such raw data with other subsidiaries.  Nonetheless, collaboration among the subsidiaries in discovering and sharing the right resources is essential for the conglomerate to succeed. Figure~\ref{fig:example} depicts a scenario of the conglomerate performing a search for resources (people, equipment, or data) based on semantic similarity across three edge locations, each controlled by an autonomous subsidiary.

\subsection{Primer}
Since the main contribution of this paper is advancing the state-of-the-art via a combination of the existing techniques of federated learning and semantic vector embedding, this section provides a brief primer on these techniques.
\subsubsection{Federated learning}
\begin{figure}[htb!]
  \includegraphics[width=\columnwidth]{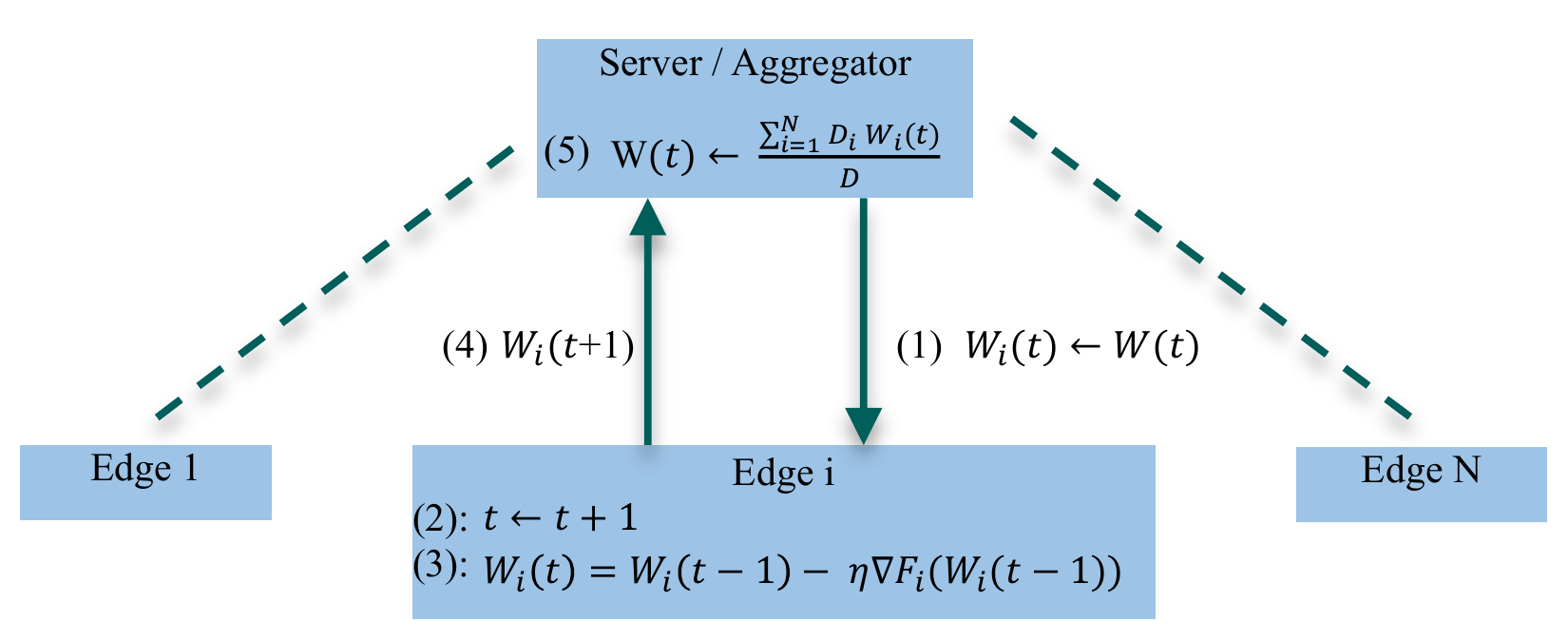}
  \caption{Federated neural-network learning via iterative federated averaging}
  \label{fig:fed_learning}
\end{figure}
In the centralized supervised neural-network learning algorithms, all training data is kept in a single environment. A loss function defines the learning objective and the weights of the neural network are adjusted via an iterative gradient descent algorithm to minimize the loss. When the data cannot be placed in a central location, like in the case of edge environments with constraints on data movement, the centralized algorithms cannot be applied. 
Figure~\ref{fig:fed_learning} depicts a distributed gradient descent with federated averaging, one of the main ideas behind federated learning. $N$ edge sites with local training data of size $D_i$ want to collaborate and jointly train a common model without sharing the raw data. A centralized server initializes the model weights at time $t$ (initially, $t=0$) $W(t)$ and distributes them to each edge site $i$. Upon receiving $W(t)$ from the server, each edge site performs a batch of local training with learning rate $\eta$ to compute local gradients $\triangledown$ that minimizes the loss function $F_i$. Based on the computed gradients $\triangledown$, each edge site updates local weights $W_i(t)$ and reports them back to the server. The server applies an aggregation function over the weight updates from all edge sites, e.g., a weighted average based on the size of the training data at each node, and distributes the aggregated weight updates back to the edge sites. This process is repeated until a predefined convergence condition is met.

\subsubsection{Doc2vec}
\begin{figure}[htb!]
  \includegraphics[width=\columnwidth]{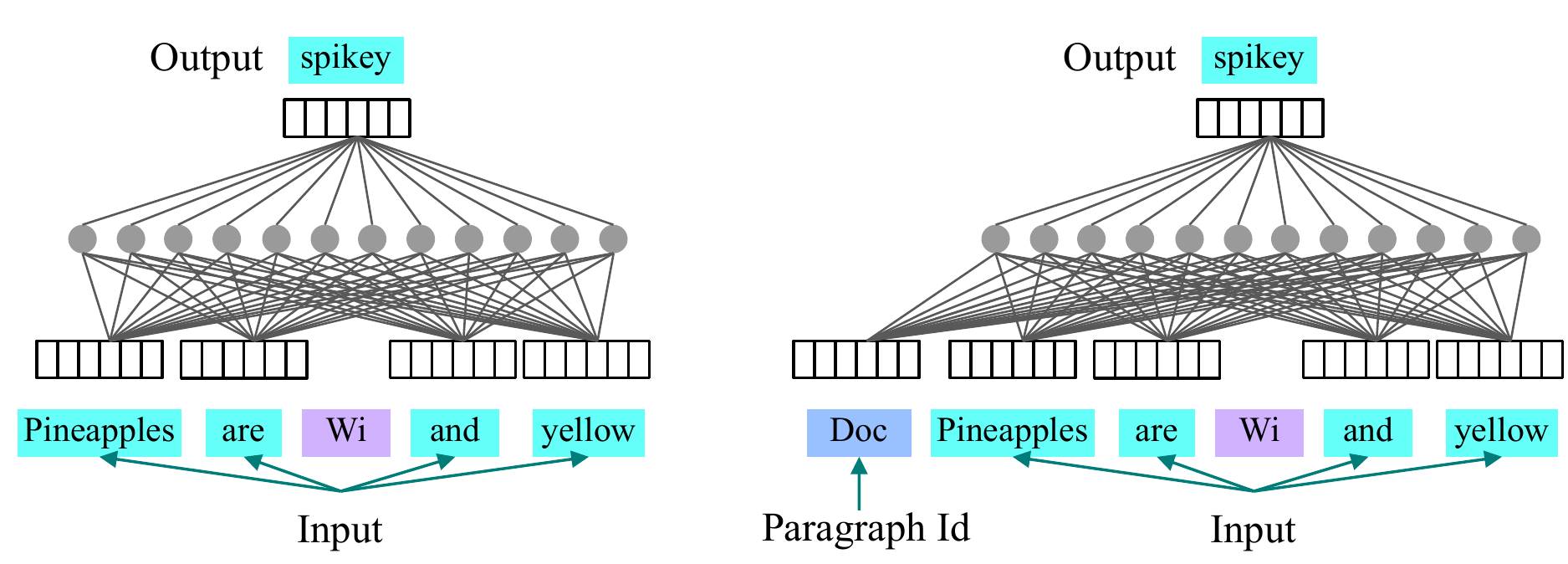}
  \caption{Left: Word2vec, Right: Doc2vec (based on CBOW algorithm (continuous bag-of-words)}
  \label{fig:doc2vec}
\end{figure}
Vector representation of text documents is useful for many purposes, e.g., document retrieval, web search, spam filtering, and topic modeling. Compared to traditional methods based on bag-of-words (BOW) and Latent Dirichlet Allocation (LDA), learned vector embedding techniques such as Doc2vec account for the context in which words and sentences are used \cite{le-doc2vec-2014}. Doc2vec is an unsupervised algorithm to learn vector embedding of text documents. The algorithm is an adaptation of Word2vec which learns vector embedding of words based on the context. 

Figure~\ref{fig:doc2vec} shows Word2vec on the left and its adaptation to Doc2vec on the right. In Word2vec, vectors of fixed dimensions representing each word in the vocabulary are initialized randomly. The learning task is defined as predicting a given word based on the preceding $N$ words and following $N$ words. The loss function is defined as the error in predicting the given word. By iterating through many sentences, the word vectors are updated to minimize the loss and accurately represent the semantic concept. Interestingly, such semantic vectors also exhibit algebraic properties, e.g., vector representing "Queen" is similar to the one corresponding to subtracting "Man" from "King" and adding "Woman". Doc2vec is a simple yet clever tweak of Word2vec where a vector representing an entire document, e.g., a paragraph, is learned along with the words in it. 

\subsubsection{Node2vec}
\begin{figure}[htb!]
  \includegraphics[width=\columnwidth]{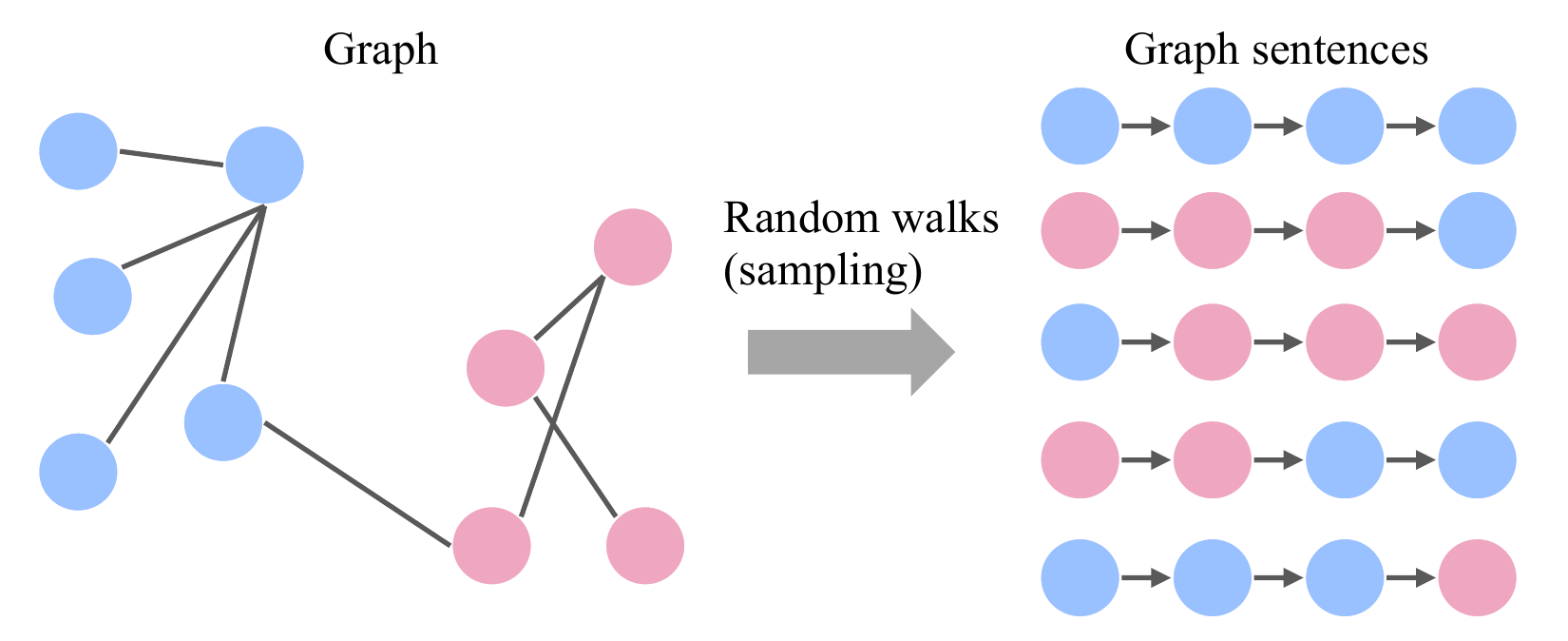}
  \caption{Node2vec: random walks on graphs}
  \label{fig:node2vec}
\end{figure}
Node2vec is an algorithm for representation learning on graph data \cite{Grover-node2vec-2016}. Given any graph, it can learn vector representations for the nodes, which can then be used for various machine learning tasks. Unlike text sentences where each word is preceded or followed by at most one word, graphs have a complex structure. The key innovation behind Node2vec is to map a graph to node sequences, a.k.a graph sentences, by generating random walks and then using Word2vec to learn the vector representation of the nodes in these sequences. Hyper parameters control number of walks to generate, walk length, and parameters to balance re-visiting previous nodes  and exploring new nodes.

\section{Problem Formulation}\label{sec:problem}
With the background on federated learning and vector embedding covered, we are ready to formally define the problem of semantic vector federation for edge environments. Important definitions and terminology are described first, followed by a formal definition of the two technical problems of joint-learning and vector-space mapping. Both technical problems aim to enable semantic search for resources across edge sites but under different assumptions. 

\subsection{Definitions}
\begin{figure}[htb!]
  \includegraphics[width=\columnwidth]{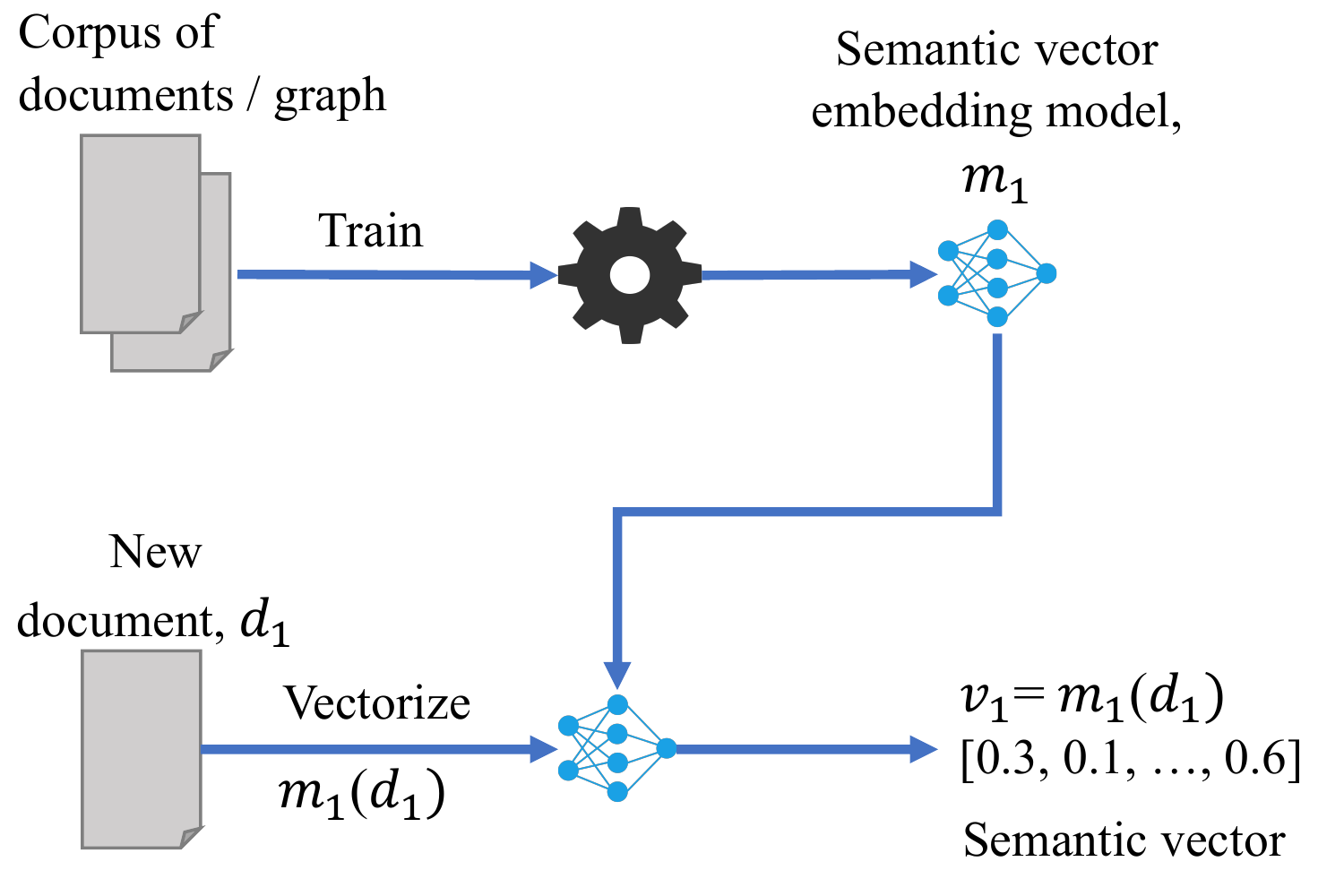}
  \caption{Definition: Semantic vector embedding}
  \label{fig:embedding}
\end{figure}

Figure~\ref{fig:embedding} shows a general definition of two key concepts in semantic vector embedding, as defined in the following.

\noindent\textbf{Semantic vector embedding model} A representation learning model $m$ trained to represent concepts as continuous (real-valued) vectors from an unlabeled dataset $D$. No assumption is made about the concept, it could be documents, graphs, expertise, consumer preferences, or other application-specific abstraction. Further, no assumption is made about the training algorithm, it could be doc2vec, node2vec, BERT, or other customized embedding algorithm. Lastly, no assumption is made about the model architecture (number of layers, neurons, input dimensions) either beyond the fact that it is a neural-network model. In the following, we adopt the functional notion of a trained model that accepts an input document and produces an output.

\noindent\textbf{Semantic vector} A continuous (real-valued) vector $v$ to represent a single unlabeled example $d$ derived as a prediction of a pre-trained semantic vector embedding model $m$ on $d$, denoted as $v = m(d)$. The dimensionality of $v$ varies and the example could be a document, a graph node, or any other application-specific abstraction. The process of generating a semantic vector is referred to as \textit{vectorization}.

\subsection{Problem: Joint-Learning}
The problem of \textit{joint-learning} of semantic vector embedding is defined as having $N$ edge sites, each edge site $i$ with local dataset $D_i$ who want to collaborate in performing global semantic similarity search given a new example $d$ across all edge sites but do not want to share their data with each other. The edge sites are willing to train a common model $m$ and distribute $m$ and $d$ to all edge sites to meet their objective.

Federated learning is a natural approach for joint-learning. However, the main challenge in applying federated learning for this purpose is in ensuring that concepts across edge sites are aligned. For example, in the case of text data if the vocabulary of words is different across edge sites, federated learning cannot be readily applied. Joint-learning assumes that beforehand, the edge sites agree on a model architecture, a training algorithm, training data input format, and semantic vector dimensionality. Hence, for many practical scenarios, joint-learning may not be viable. Also, depending on the application scenario, the example $d$ may contain sensitive information and the edge site initiating the global query may not want to share it in the raw.

\subsection{Problem: Vector-Space Mapping}
The problem of \textit{vector-space mapping} of semantic vector embedding is defined as having $N$ edge sites, each edge site $i$ with local dataset $D_i$ and a pre-trained semantic vector embedding model $m_i$ trained on $D_i$ who want to collaborate in performing global similarity search for a new example $d$ across all edge sites but do not want to share their data with each other and are not able to participate in jointly training a common model.

Vector-space mapping is a challenging problem due to two factors. First, locally trained semantic vector embedding models are trained independently of each other and hence a given example $d$ is represented as completely different semantic vectors on each edge site, so semantic similarity is not preserved across edge sites. Second, the model architectures, training algorithms, and vector dimensionality across edge sites are heterogeneous. Prior work has neither identified nor addressed these challenges. Since vector-space mapping relaxes a number of assumptions made in the case of joint learning, although non-trivial, addressing these challenges is of great significance.

\section{SEEC Algorithms}\label{sec:algo}
This section presents the SEEC algorithms for joint-learning as well as vector-space mapping along with an illustration based on the motivating example shown in Section~\ref{sec:background}. Without the loss of generality, the illustration is based on text data representing expertise of people working across edge sites, e.g., Slack conversations.

\subsection{Algorithm: Joint-Learning}
\begin{figure}[htb!]
  \includegraphics[width=\columnwidth]{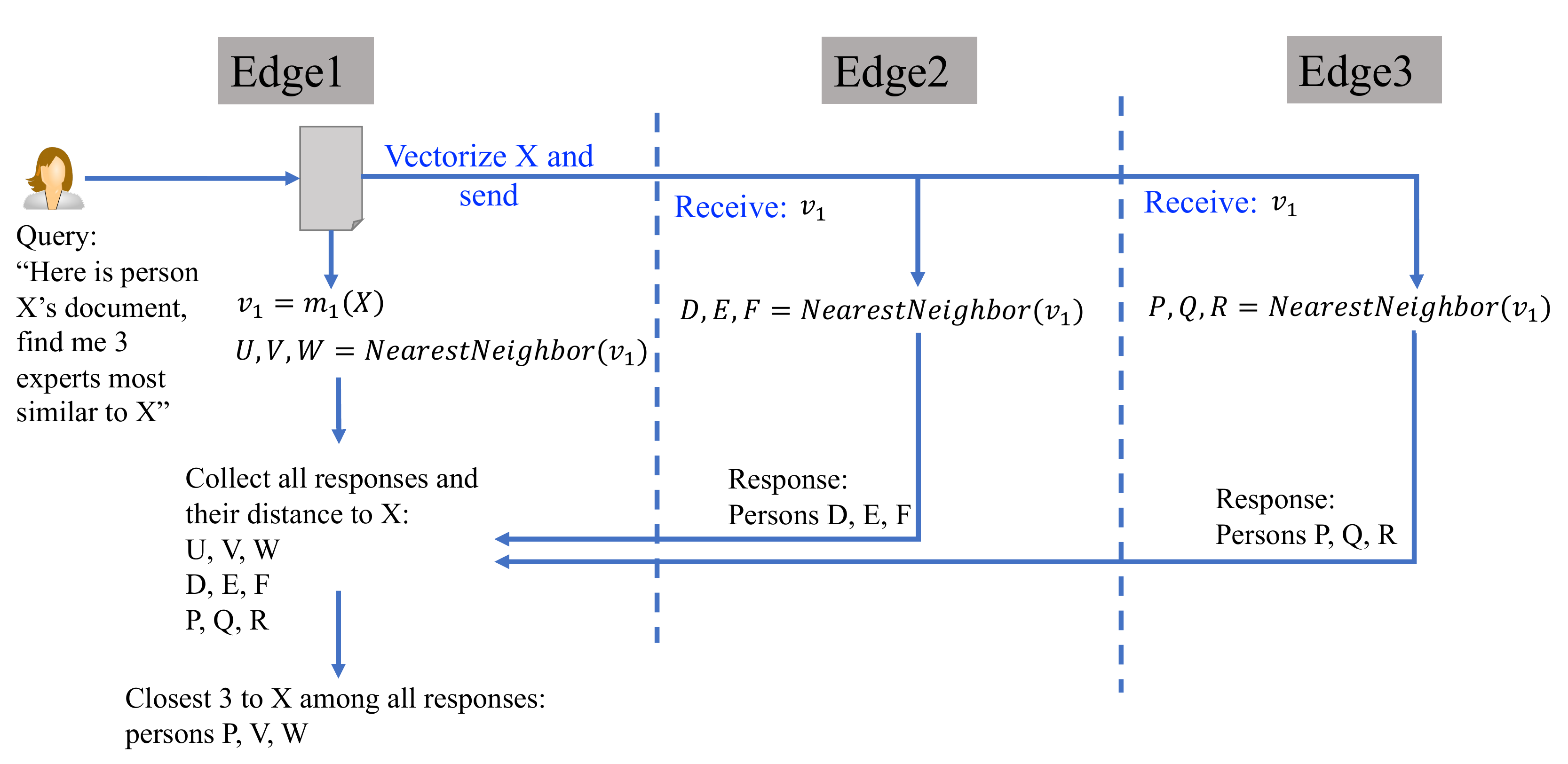}
  \caption{Semantic search for the motivating example: Joint-Learning}
  \label{fig:joint_illust}
\end{figure}

The joint-learning algorithm adapts the federated averaging algorithm \cite{mcmahan-fed-averaging-2016} to a semantic vector embedding setting. Specifically, the vocabulary of concepts must be aligned before the iterative synchronous training process to ensure consistent embedding across sites. Figure~\ref{fig:joint_illust} depicts an illustration wherein \textsc{Edge1} wants to perform a global search for top-3 experts most similar to person \textsc{X}. Assuming that a Doc2vec model $m_1$ has been distributed to all edge sites via joint-learning, \textsc{Edge1} uses $m_1$ to vectorize person \textsc{X}'s document as vector $v_1$ and sends $v_1$ to other sites. Other sites apply a similarity metric, e.g., cosine similarity, to find top-3 nearest-neighbor vectors to $v_1$ and return the corresponding person identities and cosine similarity score back to \textsc{Edge1}. After receiving the results from all edge sites, \textsc{Edge1} can select the top-3 results having the highest cosine similarity. 

A key feature of this algorithm is that person \textsc{X}'s raw document does not have to be shared with other edge sites. Further, the nearest-neighbor similarity to vectors across edge sites is valid because all vectors are produced within a single vector-space by virtue of sharing the common model $m_1$. Algorithm~\ref{algo:joint} presents the joint-learning training algorithm that is executed on the edge sites. As mentioned earlier, the main idea is similar to that of federated averaging with key innovation being the initial step of each edge site reporting their local vocabulary $U_i$ to the server and the server computing an aggregate vocabulary $U$ that is included with an initial model $m_1$, and distributing $m_1$ to all edge sites. We do not show the algorithm executed by the server as the server simply aggregates the vocabulary and distributes the initial common model, followed by averaging the weights and redistributing them in each round of training.

\begin{algorithm}[htb!]
\caption{Joint-Learning algorithm: Edge $i$}
\label{algo:joint}
\DontPrintSemicolon % Some LaTeX compilers require you to use \dontprintsemicolon instead
\SetAlgoLined
\SetKw{KwBy}{by}
\SetKwComment{Comment}{$\triangleright$\ }{}
\KwIn{Initial weights $W^0$, Local dataset $D_i$}
\KwIn{Loss function $F_i$, Local Vocabulary $U_i$}
\KwIn{Epochs $T$,  Learning rate $\eta$}
\KwOut{Updated model weights $W_i^{t}$}
 Send $U_i$ to server\;
 Receive $m_1$ from the server\;
 \For{$t \gets 0$ \KwTo $T$ \KwBy $1$}{
     Receive $W^t$ from server\;
     $W_i^t \gets W^t$\;
     $L \gets F_i (W_i^t, batch(D_i), U$ \Comment*[r]{Loss}
     $\triangledown \gets Gradient(L, F_i, W_i^t)$ \Comment*[r]{Gradient}
     $W_i^{t+1} \gets W_i^t - \eta \triangledown L$ \Comment*[r]{Weight update}
     Send $W_i^{t+1}$ to server\;
 }
 \Return{$W_i^t$}\;
\end{algorithm}

\subsection{Algorithm: Vector-Space Mapping}
A key innovation behind vector-space mapping algorithm is to train a mapper model that learns to map vectors of one vector-space to the vectors of another. Given their ability to act as universal functions, this paper uses a multi-layer perceptron (MLP) model with a single hidden layer as the mapper model architecture. However, training a MLP model requires a training dataset that is commonly available to all edge sites. Availability of such training data is highly constrained especially given that the sites do not wish to share their proprietary datasets with each other.

\begin{figure}[htb!]
  \includegraphics[width=\columnwidth]{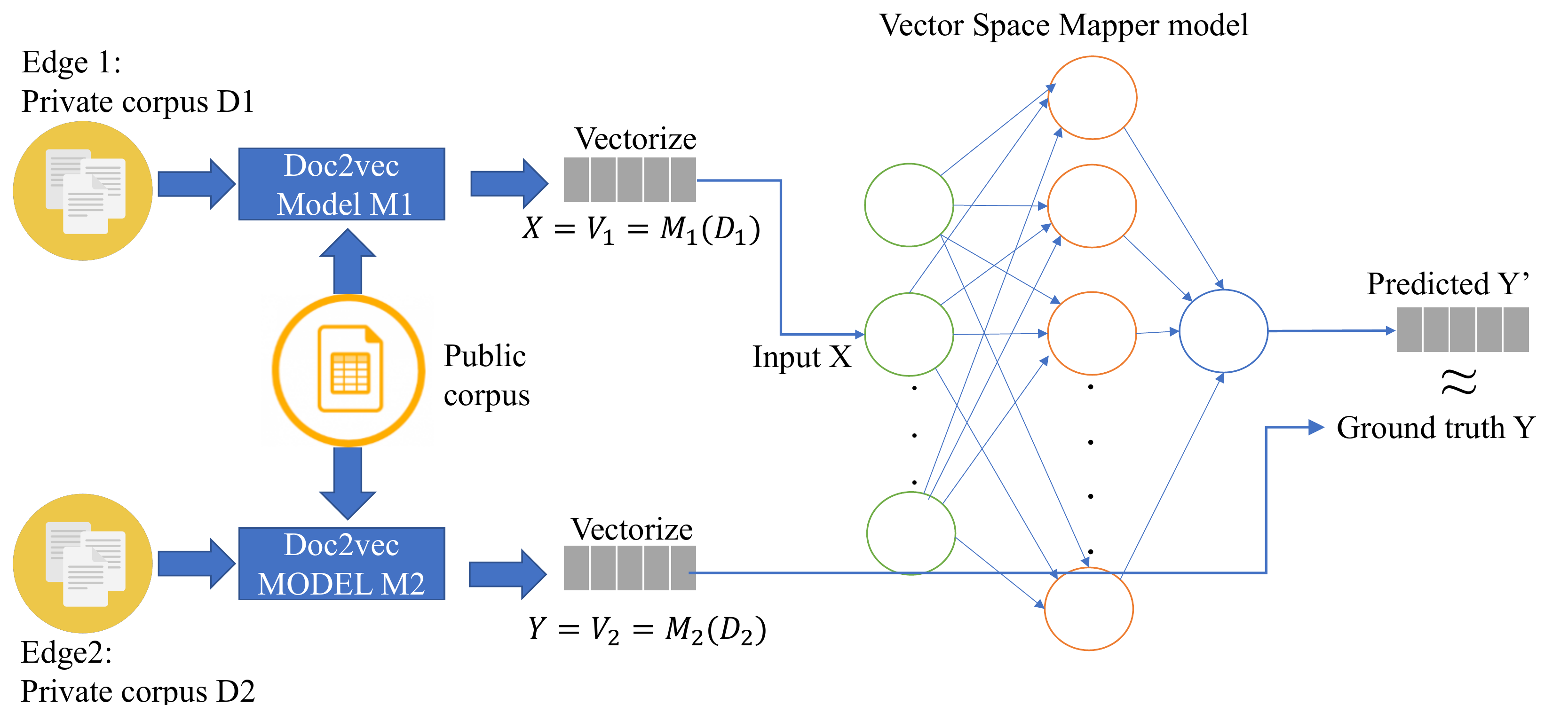}
  \caption{Learning to map vector-space of Edge1 to that of Edge2}
  \label{fig:vsm_arch}
\end{figure}

Hence, another key innovation of this paper is the idea of leveraging any publicly available corpus, regardless of its domain, as a training dataset generator for the mapper MLP model. As shown in Figure~\ref{fig:vsm_arch}, consider that a semantic vector embedding model $m_1$ is trained from local data on \textsc{Edge1} and another semantic vector embedding model $m_2$ is trained from local data on \textsc{Edge2}. The objective is to train a mapper MLP model that can map vectors produced by vector-space of $m_1$ to the vector-space of $m_2$. An auxiliary dataset $D_p$ that is accessible to both edge sites can serve as the training samples generator and facilitate the training of MLP mapper model. Input to the MLP model are the vectors produced by $m_1$ on samples of $D_p$ and the ground-truth labels are the vectors produced by $m_2$ on the same samples of $D_p$. Since the input and the output of the MLP mapper model can have different dimensionality, this approach works even when \textsc{Edge1} and \textsc{Edge2} choose different dimensionality for their semantic vectors.

\begin{figure}[htb!]
  \includegraphics[width=\columnwidth]{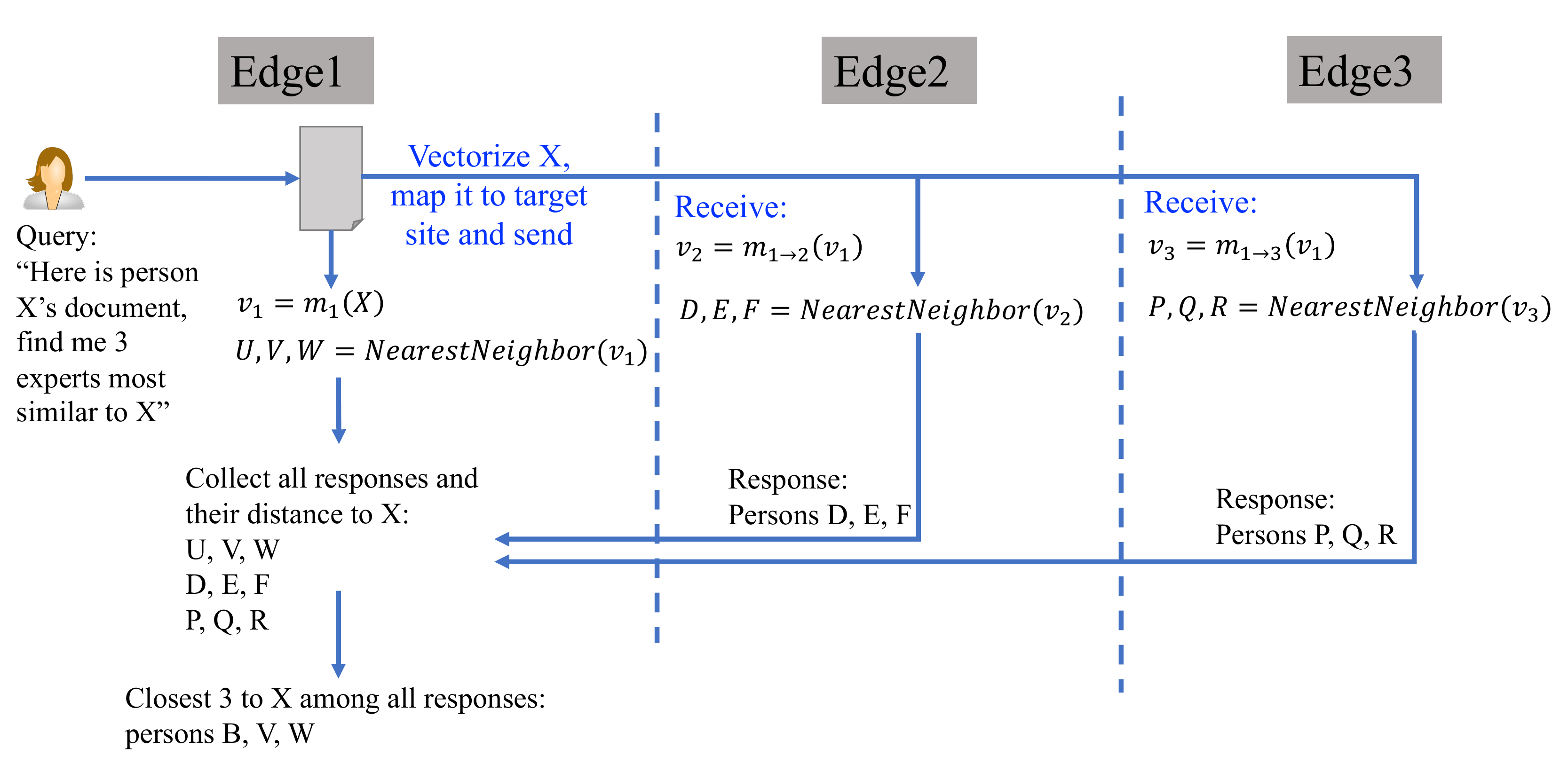}
  \caption{Semantic search for the motivating example: Vector-Space Mapping}
  \label{fig:vsm_illust}
\end{figure}

Figure~\ref{fig:vsm_illust} depicts an illustration wherein \textsc{Edge1} wants to perform a global search for the top-3 experts most similar to person \textsc{X}. Mapper models $m_{1 \rightarrow 2}$ and $m_{1 \rightarrow 3}$ have been trained to map vector-space of \textsc{Edge1} to vector-spaces of \textsc{Edge2} and \textsc{Edge3}, respectively. Unlike in joint-learning, \textsc{Edge1} uses the mapper models to map its local vector of \textsc{X} to the target edge sites \textsc{Edge2} and \textsc{Edge3}, respectively, before sending them. Algorithm~\ref{algo:vsm} presents the vector-space mapping algorithm with two components to be executed on \textsc{Edge i}: (1) Function $Map_j$ trains the mapper model $m_{i \rightarrow j}$ and (2) Function $GlobalSearch_i$ carries out global semantic search for a sample document $d$. 

\begin{algorithm}[htb!]
\caption{Vector-Space Mapping algorithm: Edge $i$}
\label{algo:vsm}
\DontPrintSemicolon % Some LaTeX compilers require you to use \dontprintsemicolon instead
\SetAlgoLined
\SetKw{KwBy}{by}
\SetKwProg{Fn}{Function}{:}{}
\SetKwComment{Comment}{$\triangleright$\ }{}
\KwIn{Local dataset $D_i$}
\KwIn{Public dataset $D_p$}
\KwIn{Loss function $F_i$}
\KwIn{Epochs $T$,  Learning rate $\eta$}
\SetKwFunction{FMain}{Main}
\Fn{\FMain{$D_i, F_i, T, \eta$}}{
     $m_i \gets TrainDoc2vec(D_i, F_i, T, \eta)$\;
     Store $m_i$ \Comment*[r]{Local model}
}
\SetKwFunction{FMain}{$Map_j$}
\Fn{\FMain{$m_j$, $D_p$, $F_i$, }}{
     $W_{i \rightarrow j} \gets RandomNN()$\;
     \ForEach{$b \in D_p$ \Comment*[r]{Batch}}{
         $v_i \gets predict(m_i, b)$ \Comment*[r]{Map input}
         $v_j \gets predict(m_j, b)$ \Comment*[r]{Map label}
         $L \gets F_i (v_i, v_j)$ \Comment*[r]{Loss}
         $\triangledown \gets Gradient(L, F_i, W_{i \rightarrow j})$\;
         $W_{i \rightarrow j} \gets W_{i \rightarrow j} - \eta \triangledown L$ \Comment*[r]{ Update}
     }
     $m_{i \rightarrow j} \gets Model(W_{i \rightarrow j}$\;
     Store $m_{i \rightarrow j}$\;
}
\SetKwFunction{FMain}{$GlobalSearch_i$}
\Fn{\FMain{$d$}}{
     \ForEach{$Edge_j \in Edges$ \Comment*[r]{Edge sites}}{ 
         $v_i \gets predict(m_i, d)$ \Comment*[r]{Vectorize}
         $v_j \gets m_{i \rightarrow j}(v_i)$ \Comment*[r]{Map to site j}
         Send query $v_j$ to $Edge_j$\;
         $V_{sim} \gets$ Receive result vectors from $Edge_j$\;
     }
     \Return $V_{sim}$\;
}
\end{algorithm}

\section{Experiments}\label{sec:exp}
We evaluate the two algorithms of joint-learning and vector-space mapping via extensive experiments on two data modalities: natural language and graph. The experiments are anchored on the motivating example described in Section~\ref{sec:motivation} with the task of performing a global semantic search for individuals with expertise. The evaluation approach is three-pronged. First, an objective evaluation of how well the SEEC algorithms perform relative to the baseline of centralized models. Comparing with a baseline is a standard practice for unsupervised algorithms since there is no ground truth on semantic similarity between samples. Second, an objective evaluation compared to a small number of ground truth samples collected via a user survey. And third, a subjective examination of the semantic search results produced by the two SEEC algorithms, joint-learning and vector-space mapping. 

\subsection{Experimental Setup}
Before we discuss the experimental results, we describe the dataset, the implementation details, and the metrics used for the objective evaluation.

\subsubsection{Datasets}
For the natural language modality, we leverage three different datasets: (a) an IBM internal dataset consisting of Slack collaboration conversations, (b) the 2017 Wikipedia dataset with 100K samples \cite{Wikipedia-2017}, and (c) the 20-newsgroup public dataset with $18846$ samples \cite{Lang95}. For joint-learning experiments (a) is used for both the centralized and federated experiments. For vector-space mapping experiments, both (a) and (b) are used as private datasets with (c) acting as the public dataset accessible by all edge sites. For the graph modality, we leverage (a) above for the joint-learning experiments but instead of looking at the text content of the posts, we construct a collaboration graph between users. 

The Slack dataset (a) consists of natural language conversations across $7367$ slack channels among $14208$ unique users. Of these, only $1576$ users having sufficient activity (more than 100 posts) are used in the experiments. All Slack posts of a user is treated as a single document in training the Doc2vec models. For the centralized case, Slack posts of all users are used for training a single Doc2vec model whereas for the federated case (joint-learning and vector-space mapping) the users are uniformly distributed across two edge sites. No additional knowledge of the organization hierarchy, projects, or teams is included, leaving the models to rely solely on the content of the Slack posts as a basis of semantic vector embedding representing each user.

In constructing a graph from the Slack dataset, each user is treated as a node in the graph and other users that participate in the same Slack channel as the user are treated as the edges. For avoiding noisy edges due to having channels with a large number of users, a pair of users participating together, i.e., co-occurring, in less than $10$ channels do not have an edge between them. Another approach would have been to assign weights to edges, however, Node2vec does not take advantage of edge weight information. The entire graph is used for training the centralized Nodes2vec model. For the federated case, users are randomly assigned to one of the Edge sites. When doing so the cross-site edges are handled in two alternative ways: (1) the cross-sites edges are not retained, so each edge site has edges only among the users assigned to the site, called \textit{no retention}, and (2) the nodes involved on cross-site edges are retained on both sites, called \text{retention}. 

Beyond the above datasets, we surveyed $36$ IBM employees and asked them to identify top-3 collaborators who possess the most similar skills as them (in no particular order). The precise question that was asked was: \textit{Name the 3 people in your company who in your judgment possess skills most similar to your own}. We collected 36 responses, of which $9$ of them were from individuals who had significant Slack activity (over $100$ posts). Although this is a small number of samples, we treat the survey results from these $9$ employees as ground-truth on semantic similarity.

\subsubsection{Implementation}
We leverage the gensim \cite{Gensim-2010} framework for training semantic vector embedding models as well as performing similarity search. The similarity measure used for nearest-neighbor calculations in semantic search is cosine-similarity, to allow for meaningful measurement in the case of high-dimensional vectors. For the natural language dataset, we use the Doc2vec model architecture with the skip-gram PV-DM algorithm with $40$ epochs and learning rate of $0.025$. Doc2vec semantic vectors are 50-dimensional real-valued vectors. For the graph dataset, we use the Node2vec architecture with $40$ epochs and learning rate of $0.025$. Node2vec semantic vectors are 124-dimensional real-valued vectors. The hyper-parameters of the Node2vec favor homophily approach where the return parameter $p$ is favored over the in-out parameter $q$. We set $p=0.6$ and $q=0.1$. The walk length parameter, the number of hops to other nodes from the start node is set to $20$ and the number of walks, the number of iterations of node-hopping to perform, is also set to $20$. 

In the case of vector-space mapping, the mapper model is an MLP model with a single hidden layer with $1200$ neurons and a dropout ratio of $0.2$. We use the cosine embedding loss in training the MLP as the semantic similarity is based on cosine similarity. \textsc{Adam} optimizer with a learning rate of $0.00001$ and $20$ epochs of training with the batch size of $64$ were applied. 

It is worth emphasizing that these details are provided for completeness and these parameters are quite commonly used in the literature.  The objective in this paper is not to have the best-performing semantic vector embedding models. Instead, we're primarily interested in evaluating the \textit{relative} performance of the federated algorithms compared to the traditional centralized ones. Hence, all of the above parameters are kept the same for centralized and the federated case.

\subsubsection{Metrics}
For objectively measuring how well the federated algorithms perform relative to the centralized case without the survey data, the degree of overlap is computed as follows. For a given document $d$ in the dataset, the centralized model is used to vectorize the document and the set of top-k most similar documents from the dataset are found based on cosine-similarity, called $d_{c}^{k}$. Then, using the respective federated algorithm, the set of top-k most similar documents are found for the same document $d$, called $d_{f}^{k}$. The degree of overlap $sim_{k}$ then is the ratio of cardinality of the intersecting set and $k$, denoted as  $sim_{k} = \frac{|d_{c}^{k} \cap d_{f}^{k}|}{k}$. For multiple documents in the dataset, a simple mean of $sim_{k}$ is computed over all documents. When $sim_{k}=1$, the centralized and the federated model produce identical results on semantic search. The idea behind the measure is simple: the higher the $sim_{k}$, the closer the federated case performance is to the centralized case. In evaluating the federated algorithms relative to the centralized case we set $k=10$. In evaluating the federated algorithms relative to the survey results, we set $k=3$ as the employees were asked to identify top-3 most similar employees. 

For objectively measuring how well the federated algorithms perform relative to the centralized case with the survey data, $sim_k$ is computed between the centralized case and the survey result as well as between the federated case and the survey result, for each participant. Then, a Pearson's correlation between the $sim_k$ values is reported. The correlation between $sim_k$ is a better measure than raw $sim_k$ because we are interested in the similarity between the overlaps achieved by the centralized case and the federated case, rather than the actual overlap with the survey results themselves. The raw $sim_k$ itself is largely dependent on whether or not the participants used Slack in the same way to collaborate.

\subsection{Results: Joint-Learning}
\subsubsection{Natural language}
\begin{figure}[htb!]
  \includegraphics[width=\columnwidth]{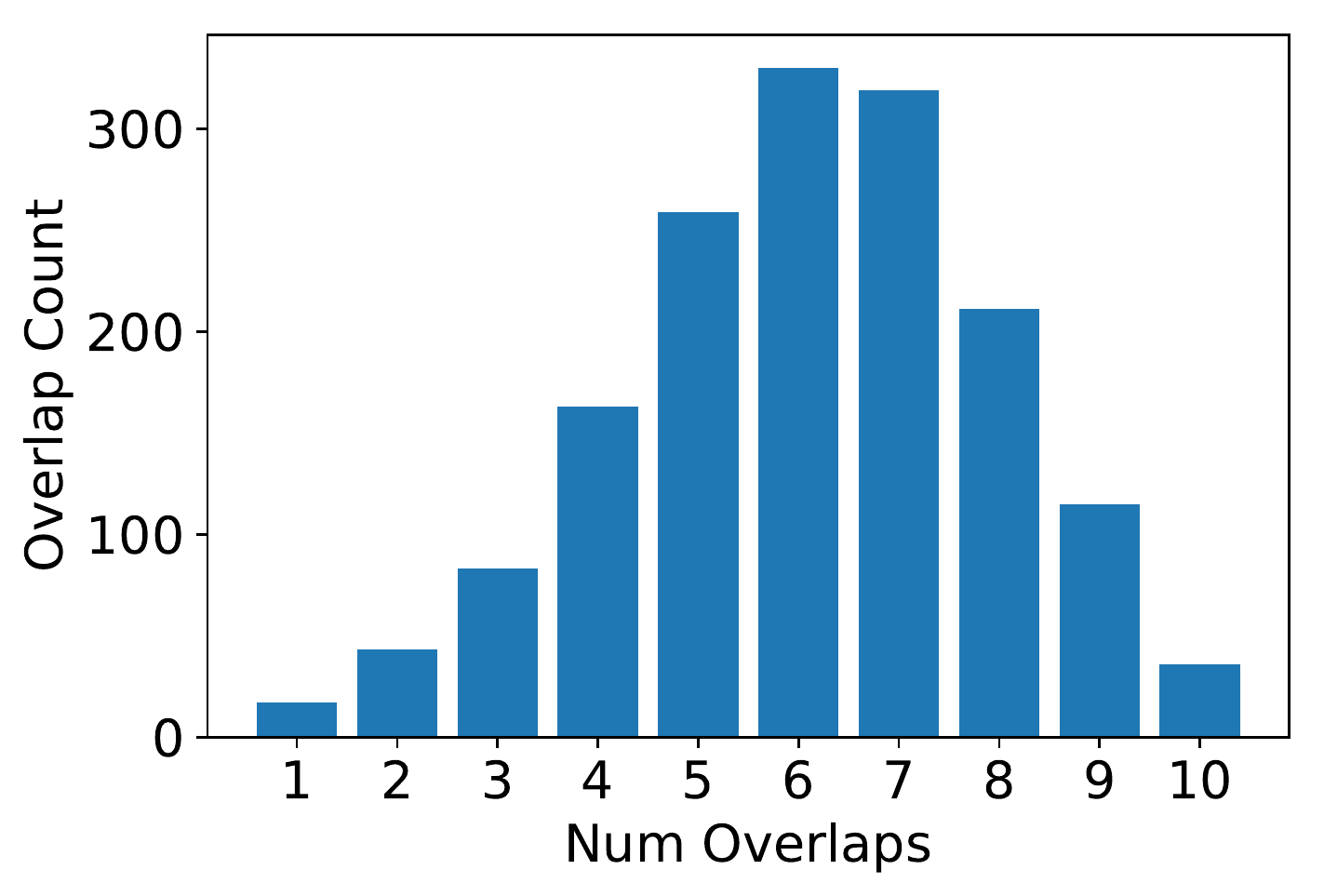}
  \caption{Performance of Doc2vec joint-learning relative to centralized learning, $sim_{k}=0.609$}
  \label{fig:fed_doc2vec_centralized}
\end{figure}

Figure~\ref{fig:fed_doc2vec_centralized} shows the distribution of the number of overlaps between the centralized case and the joint-learning case ($sim_{10} \times 10$) when the joint-learning algorithm is applied to the Slack dataset. As indicated by the $sim_{k}$ of $0.609$, for majority of the users, the joint-learning model found about $6$ of the same users found by the centralized model.

\begin{figure}[htb!]
  \includegraphics[width=\columnwidth]{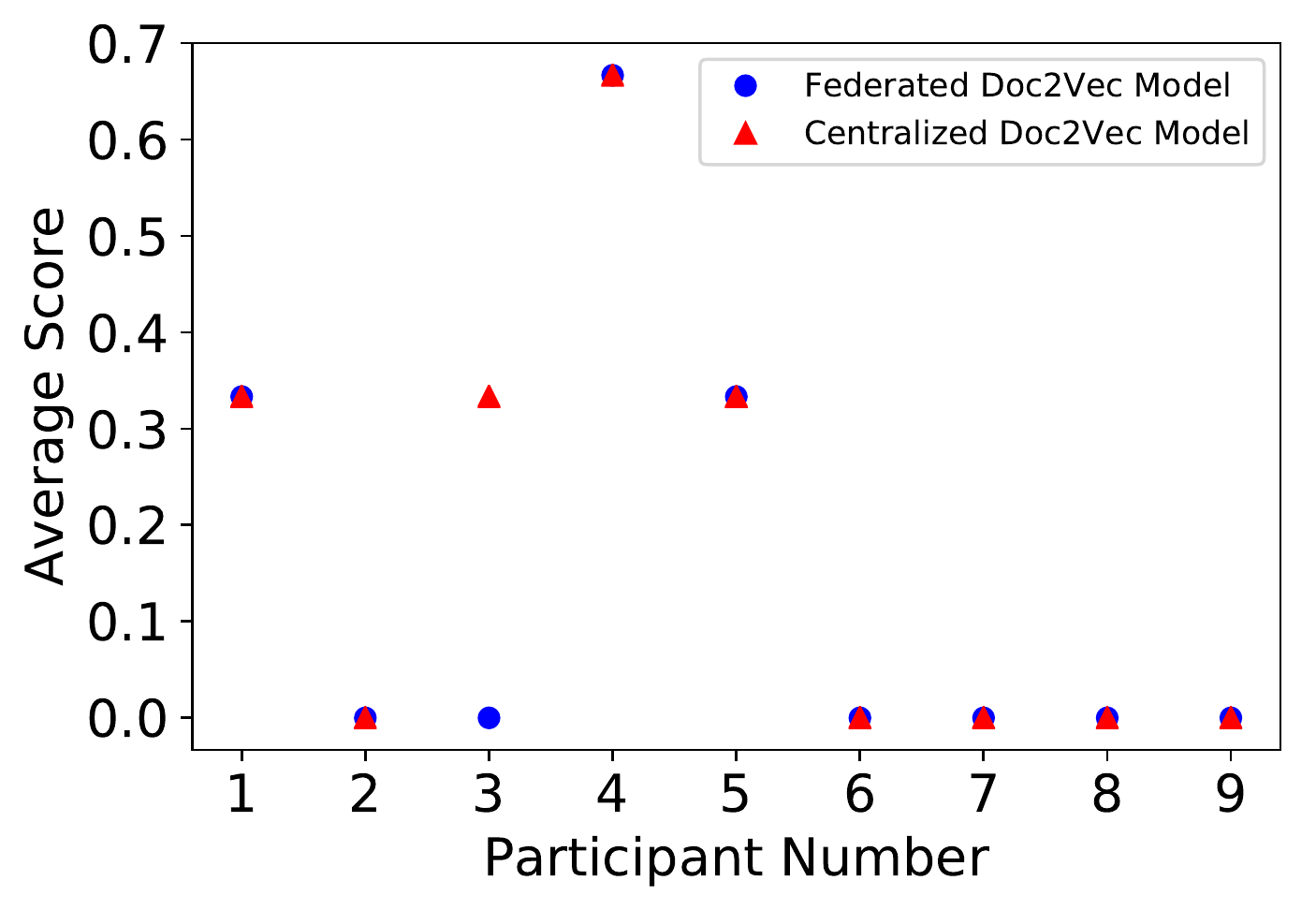}
  \caption{Performance of Doc2vec joint-learning relative to the survey}
  \label{fig:fed_doc2vec_survey}
\end{figure}

Figure~\ref{fig:fed_doc2vec_survey} shows the $sim_k$ score corresponding to the centralized Doc2vec model relative to the survey and the same corresponding to the joint-learning Doc2vec model relative to the survey, for each participant in the survey. Barring participant 3, the centralized and the joint-learning models produce the same $sim_k$ score, which is promising. The Pearson's correlation between the $sim_k$ of the centralized and the joint-learning models in this case was $0.89$, which indicates a clear similarity in their performance.

Based on the above, we can conclude that there is not a significant loss in accuracy introduced by the joint-learning algorithm, when compared to the centralized model. Thus, the joint-learning Doc2vec model and centralized Doc2vec model have similar performance, making the joint-learning algorithm a viable alternative to the centralized case. One important thing to note is that both models were able to score matches on at least a third of the participants in the survey. Since the probability of getting a match when choosing randomly would be almost zero, the fact that the models were able to produce some matches shows that both models were indeed able to learn key features about the user from their Slack posts.

\subsubsection{Graph}
\begin{figure}[htb!]
  \includegraphics[width=\columnwidth]{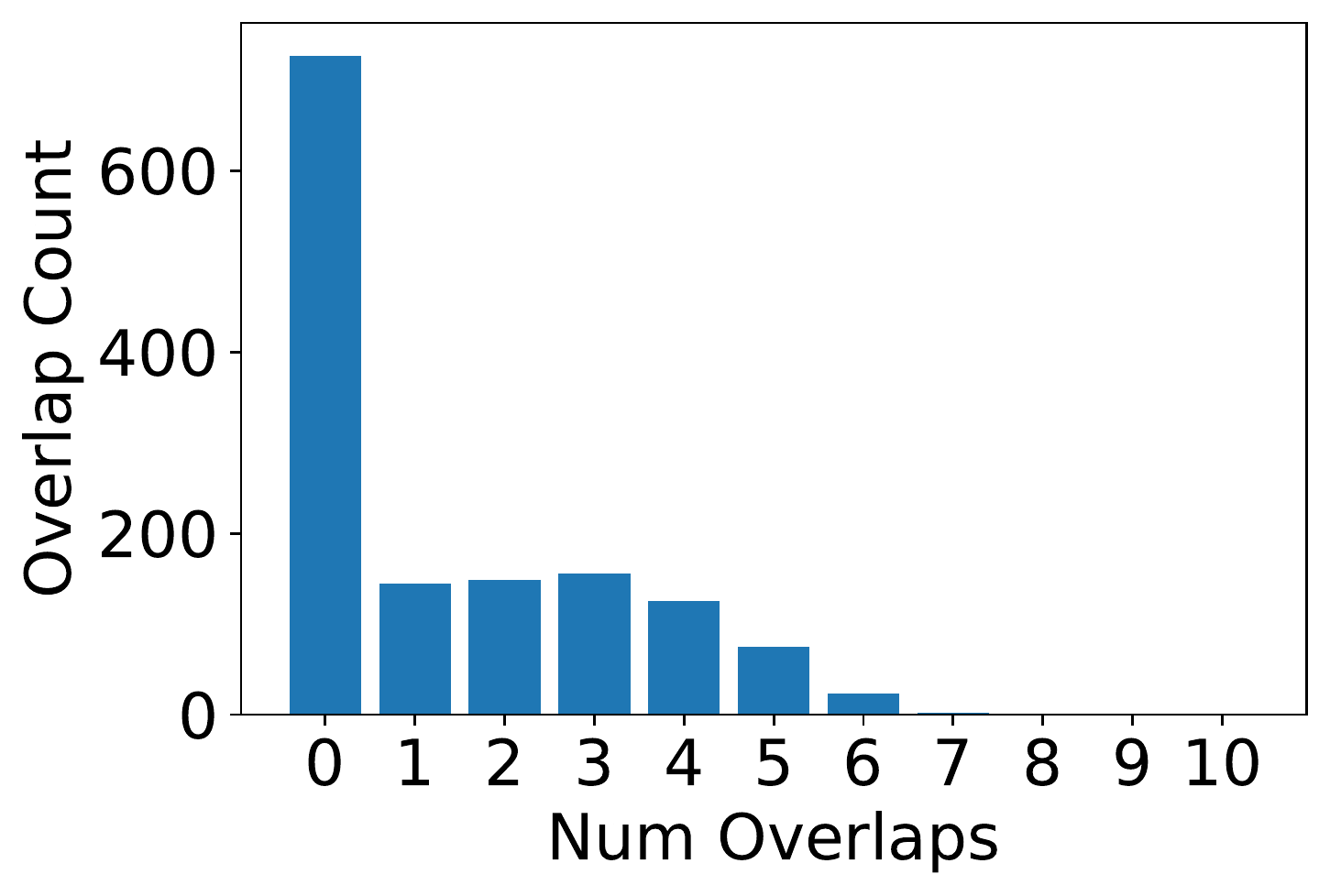}
  \caption{Performance of Node2vec joint-learning relative to centralized learning, no retention, $sim_{k}=0.138$}
  \label{fig:fed_node2vec_centralized}
\end{figure}

Figure~\ref{fig:fed_node2vec_centralized} shows the distribution of the number of overlaps between the centralized case and the joint-learning case ($sim_{10} \times 10$) when the joint-learning algorithm is applied to the graph dataset with no retention of cross-site collaborators. As seen in the distribution as well as indicated by the $sim_{k}$ of $0.138$, for majority of the users, the joint-learning model found almost no users returned by the centralized model. This is not an encouraging result by itself. However, since the cross-site edges are dropped from the graph corresponding to the joint-learning case, valuable information about those users' collaboration behavior is lost compared to the centralized case having the entire graph. Although this explanation is intuitive, to validate it, we need to examine the result when the cross-site collaborators are retained, discussed next.

\begin{figure}[htb!]
  \includegraphics[width=\columnwidth]{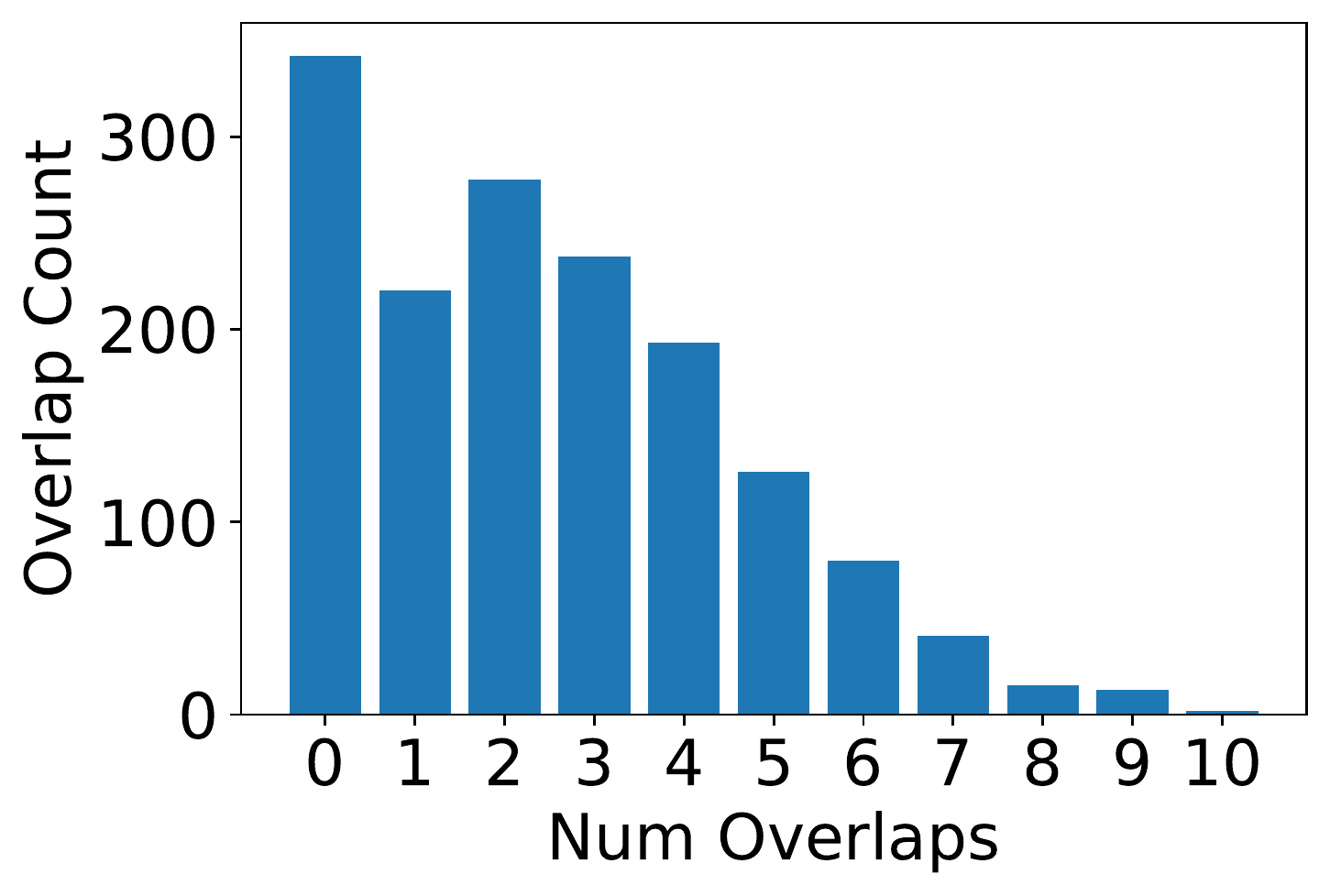}
  \caption{Performance of Node2vec joint-learning with collaborator retention relative to centralized learning, $sim_{k}=0.253$}
  \label{fig:fed_node2vec_copy_centralized}
\end{figure}

Figure~\ref{fig:fed_node2vec_copy_centralized} shows the distribution of the number of overlaps between the centralized case and the joint-learning case ($sim_{10} \times 10$) when the joint-learning algorithm is applied to the graph dataset with all cross-site collaborators retained across both sites. As seen in the distribution as well as indicated by the $sim_{k}$ of $0.253$, for majority of the users, the joint-learning model found more than $2$ of the same users returned by the centralized model. Compared to the no retention result, this is a significantly better result. Thus, the explanation above is validated as retaining the cross-site collaborators clearly helps the joint-learning model achieve more accurate user embedding. 

Although the difference between the Node2vec joint-learning results above can be explained by the difference in retention policy, the inferior results of Node2vec when compared with Doc2vec requires further investigation. One hypothesis is that the random assignment of users to edge sites can have an adverse effect on the joint-learning performance because such an assignment can have an uneven effect on the collaborative user clusters in the graph. For example, one edge site may end up having most of its collaborative clusters unaffected whereas another may have it's collaborative clusters split into two sites.  Although the immediate collaborators may be preserved via cross-site retention, the higher-order collaborations are still affected.

This hypothesis was investigated via a subjective examination of the semantic search results and found to be accurate in many cases.  The results in Table~\ref{tab:joint_node2vec} show one such example for a given user. The first column shows the top-10 most similar users returned by the centralized Node2vec model for the given user. The second and the third columns show the top-10 most similar users returned by the joint-learned Node2vec model from each edge site with cross-site collaborators retained. Each row is a distinct user to clearly show overlapping results. Clearly, the overlap between the first edge site and the centralized model is much greater ($9$ users) than the overlap between the second edge site and the centralized model ($5$ users). Such an asymmetry when the number of users across sites is the same implies an asymmetry in the connectedness of the graph across sites.

\begin{table}
\centering
\small
\begin{tabular}{c|c|c|c}
\hline
User & Centralized & \multicolumn{2}{c}{Joint-Learning}\\
&  & Edge1 & Edge2\\
\hline
User1 & Yes & Yes & \\
User2 & Yes & Yes & Yes \\
User3 & Yes & Yes & Yes \\
User4 & Yes & Yes & Yes \\
User5 & Yes & Yes & Yes \\
User6 &     &     &     \\
User7 & Yes & Yes & Yes \\
User8 & Yes & Yes &     \\
User9 & Yes & Yes &     \\
User10 & Yes & Yes &    \\
User11 &     & Yes &    \\
User12 &    &     & Yes \\
User13 &    &     & Yes \\
User14 &    &     & Yes \\
User15 &    &     & Yes \\
User16 &    &     & Yes \\
\hline
\end{tabular}
\caption{Semantic search on Joint-Learning Node2vec trained on Slack data with collaborator retention across edge sites}
\label{tab:joint_node2vec}
\end{table}

%User1: kolodner
%User2: sima
%User3: roee.shlomo
%User4: tomer.solomon
%User5: mayaa
%User6: shelly
%User7: etyk
%User8: lanna
%User9: eranra
%User10: David.Ohana
%User11: brunow
%User12: j.lynch
%User13: rovallam
%User14: rodsbandarra
%User15: rchebabi
%User16: hillelm

%
\begin{figure}[htb!]
  \includegraphics[width=\columnwidth]{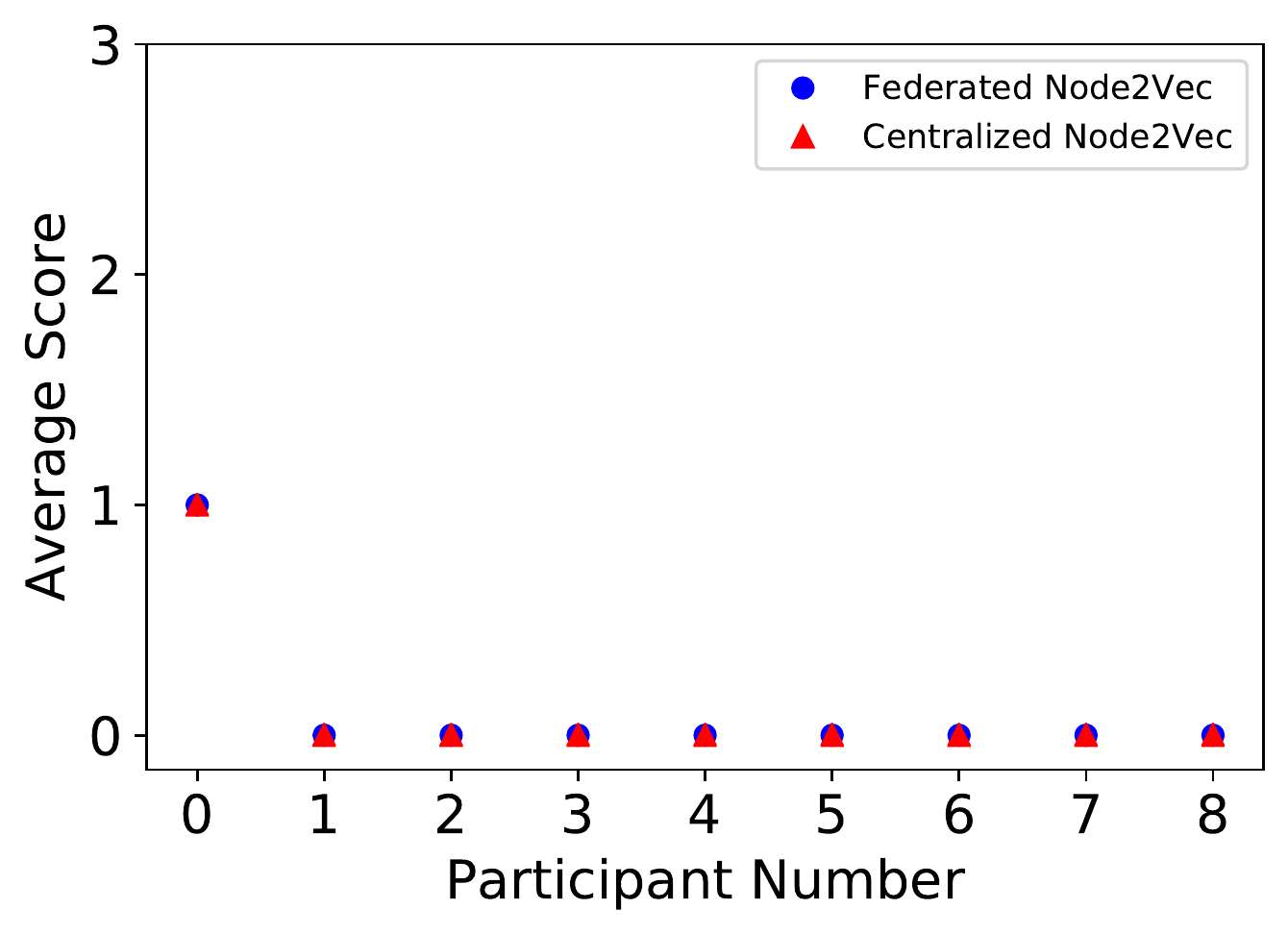}
  \caption{Performance of Node2vec joint-learning, no retention, relative to a survey}
  \label{fig:fed_node2vec_no_copy_survey}
\end{figure}

Figure~\ref{fig:fed_node2vec_no_copy_survey} shows the number of overlaps corresponding to the centralized Node2vec model relative to the survey and the same corresponding to the joint-learning Node2vec model (no retention) relative to the survey, for each participant in the survey. The centralized and the joint-learning models produce the same number of overlaps relative to the survey, which is promising. The Pearson's correlation between the $sim_k$ of the centralized and the joint-learning models in this case is $1.0$, which indicates a clear similarity in their performance.

\begin{figure}[htb!]
  \includegraphics[width=\columnwidth]{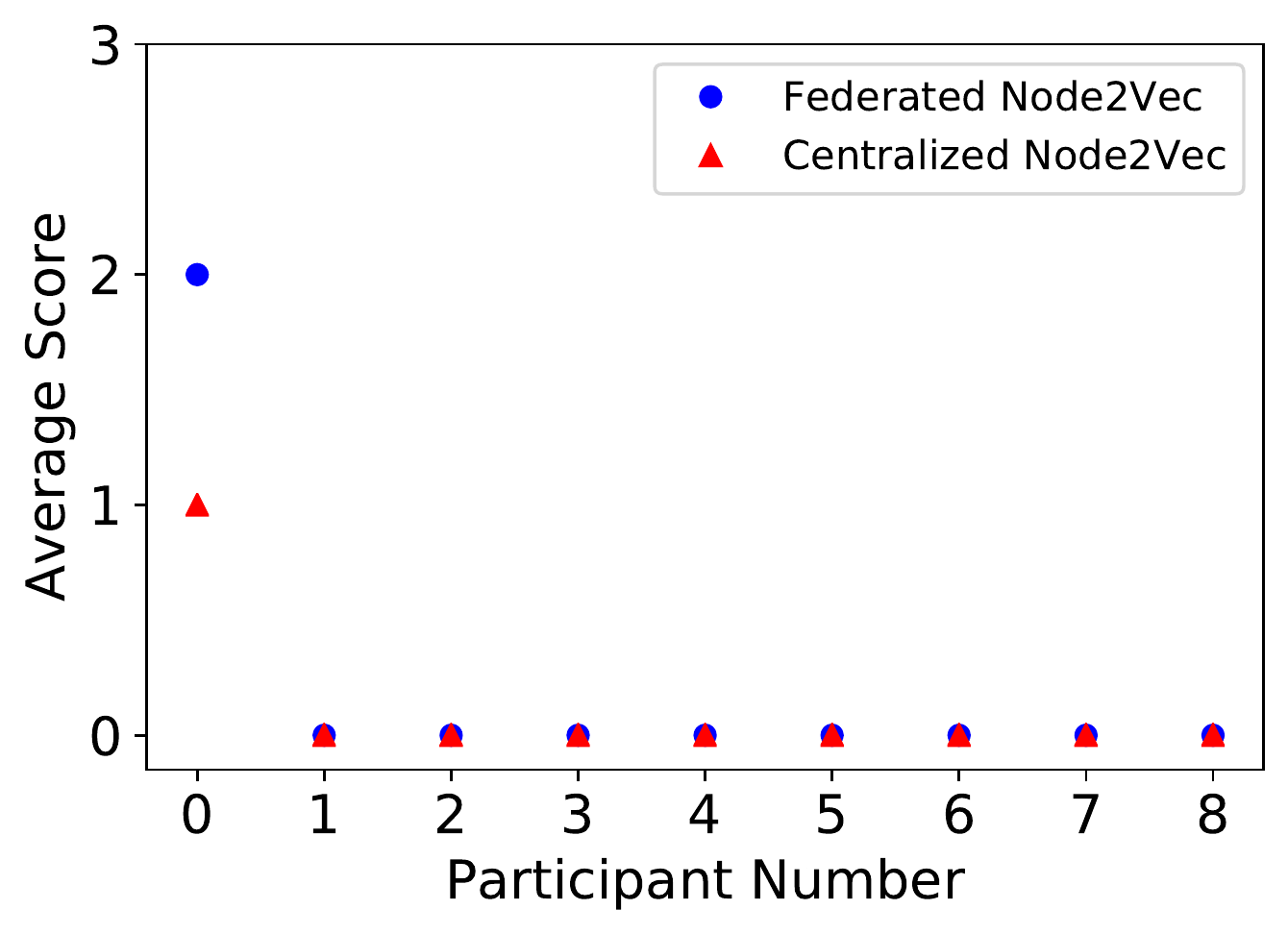}
  \caption{Performance of Node2vec joint-learning, with collaborator retention, relative to a survey}
  \label{fig:fed_node2vec_copy_survey}
\end{figure}

Similarly, Figure~\ref{fig:fed_node2vec_no_copy_survey} shows the number of overlaps corresponding to the centralized Node2vec model relative to the survey and the same corresponding to the joint-learning Node2vec model (with collaborator retention) relative to the survey, for each participant in the survey. Except for the first participant, the centralized and the joint-learning models produce the same number of overlaps relative to the survey, which is promising. Interestingly, the joint-learned Node2vec model appears to reinforce certain clusters and thus returns higher overlap with the survey than the centralized model.  This may suggest that a positive side-effect of partitioning graph across sites may be to drop extraneous and noisy edges, helping the edge site graph be more accurate in some cases.

Overall, the results suggest there is value in the joint-learning algorithm for graph datasets, although the results are not as promising as with the natural language dataset. The algorithm does identify several clusters of nodes which correspond to teams. Perhaps the channel co-occurrence is a poor feature for modeling users. As stated above, the Node2Vec model generated for all cases was homophily-focused in nature rather than structurally equivalent. It seeks to identify highly connected nodes (people who work together, are on the same team) rather than nodes of similar structure (people acting in a similar role, i.e., a manager). 

\subsection{Results: vector-space Mapping}
First we illustrate the impact of not having any mapping across vector-spaces with an example from the Wikipedia dataset. Then, we present the vector-space mapping results with two different datasets that reveal an interesting dependence on the richness of the vocabulary. First, we use the Slack data as the private data on each edge site and the 20-newsgroup data as the public dataset for training the mapper model. Next, we use the Wikipedia dataset as the private data on each edge site and the 20-newsgroup data as the public dataset for training the mapper model. 

\subsubsection{No mapping}
To illustrate the contrast between unmapped vector-spaces and the ones with mapping, we select two Wikipedia articles from $m_1$ for examination on \textit{Anarchism} and on \textit{Nintendo}, and performed a similarity search on each of the two articles on the centralized Doc2vec model.

\begin{table}
\centering
\small
\begin{tabular}{lc|lc}
\hline
\multicolumn{2}{c|}{Anarchism}  & \multicolumn{2}{c}{Nintendo}\\
Article & Score & Article & Score\\
\hline
Libertarian socialism & 0.94 & PlayStation (console) & 0.92\\ 
Anarcho-communism & 0.91 & GameCube & 0.91\\
Individualist anarchism & 0.90 & Handheld game console & 0.91\\ 
Anarcho-pacifism & 0.84 & SNES & 0.90\\
The ego and its own & 0.84 & PlayStation3 & 0.90\\
\hline
\end{tabular}
\caption{Semantic search on centralized Doc2vec model trained on Wikipedia data}
\label{tab:vsm_centralized}
\end{table}

Table~\ref{tab:vsm_centralized} shows the top-5 most similar articles to "Anarchism" and "Nintendo" based on their cosine-similarity score. The articles clear show a significant degree of semantic similarity with the respective query articles, which is as expected. We shall refer to this result to compare the case where the same search is performed across the two sites without any vector-space mapping and then when it is repeated but a mapper model is used to map the query vector before sending it to other sites.

In the no mapping case, the first step is to vectorize ‘Anarchism’ article on \textsc{Edge1} using the local model $m_1$, producing $v_1$. Next, $v_1$ is sent to \textsc{Edge2} as-is and top-5 most similar article to $v_1$ are selected on both edge sites. In the mapping case, the first step is to vectorize ‘Anarchism’ article on \textsc{Edge1} using the local model $m_1$, producing $v_1$. Next, $v_1$ is mapped via the mapper model $m_{1 \rightarrow 2}$ to the vector-space of \textsc{Edge2} as $v_2$ and sent to \textsc{Edge2}. From \textsc{Edge1}, top-5 most similar articles to $v_1$ and from \textsc{Edge2}, top-5 most similar articles to $v_2$ are returned.

\begin{table*}
\centering
\small
\begin{tabular}{lc|lc|lc}
\hline
\multicolumn{2}{c|}{Anarchism: Edge1}  & \multicolumn{2}{c|}{Anarchism: Mapped to Edge2} & \multicolumn{2}{c}{Anarchism: Edge2, no mapping} \\
Article & Score & Article & Score & Article & Score\\
\hline
Anarcho-pacifism & 0.84 & Libertarian socialism & 0.87 & The Ego and Its Own & 0.51\\ 
Socialism & 0.84 & Anarcho-communism & 0.83 & The Closing of the American Mind & 0.51\\
Anarcho-syndicalism & 0.83 & Individualist anarchism & 0.83 & Treason & 0.50\\
Anarchism and violence & 0.83 & Spanish Revolution of 1936 & 0.82 & Giuseppe Mazzini & 0.50\\
Ideology & 0.83 & Louis Althusser & 0.81 & Organized crime & 0.49\\
\hline
\hline
\multicolumn{2}{c|}{Nintendo: Edge1}  & \multicolumn{2}{c|}{Nintendo: Mapped to Edge2} & \multicolumn{2}{c}{Nintendo: Edge2, no mapping} \\
Article & Score & Article & Score & Article & Score\\
\hline
GameCube & 0.94 & PlayStation3 & 0.92 & Lester Patrick Trophy & 0.55\\
Nintendo 64 & 0.92 & Game Gear & 0.89 & Art Ross Trophy & 0.52\\
Virtual Boy & 0.90 & PlayStation (console) & 0.89 & Elo rating system & 0.51\\
Master System & 0.89 & Video game console & 0.88 & Duck Hunt & 0.50\\
3DO Interactive Multiplayer & 0.89 & SNES & 0.87 & Cheating in poker & 0.50\\
\hline
\end{tabular}
\caption{Semantic search with vector-space mapping of Doc2vec model trained on Wikipedia data}
\label{tab:vsm_federated}
\end{table*}

Table~\ref{tab:vsm_federated} shows the comparison of the mapping and the no mapping case. The first column shows the top-5 most similar results on \textsc{Edge1} using local model $m_1$. The second columns shows the top-5 most similar results on \textsc{Edge2} with mapping whereas the third column shows the top-5 most similar results without mapping.

Examining the third column, clearly, without vector-space mapping, the results bear no semantic similarity with the respective query article on anarchism or Nintendo. The second column on the other hand shows significant similarity with the query article as well as with the centralized model results shown in Table~\ref{tab:vsm_centralized}. Thus, both results illustrate the effectiveness of vector-space mapping algorithm for semantic search across independently trained local models.

\subsubsection{Slack as private data}
\begin{figure}[htb!]
  \includegraphics[width=\columnwidth]{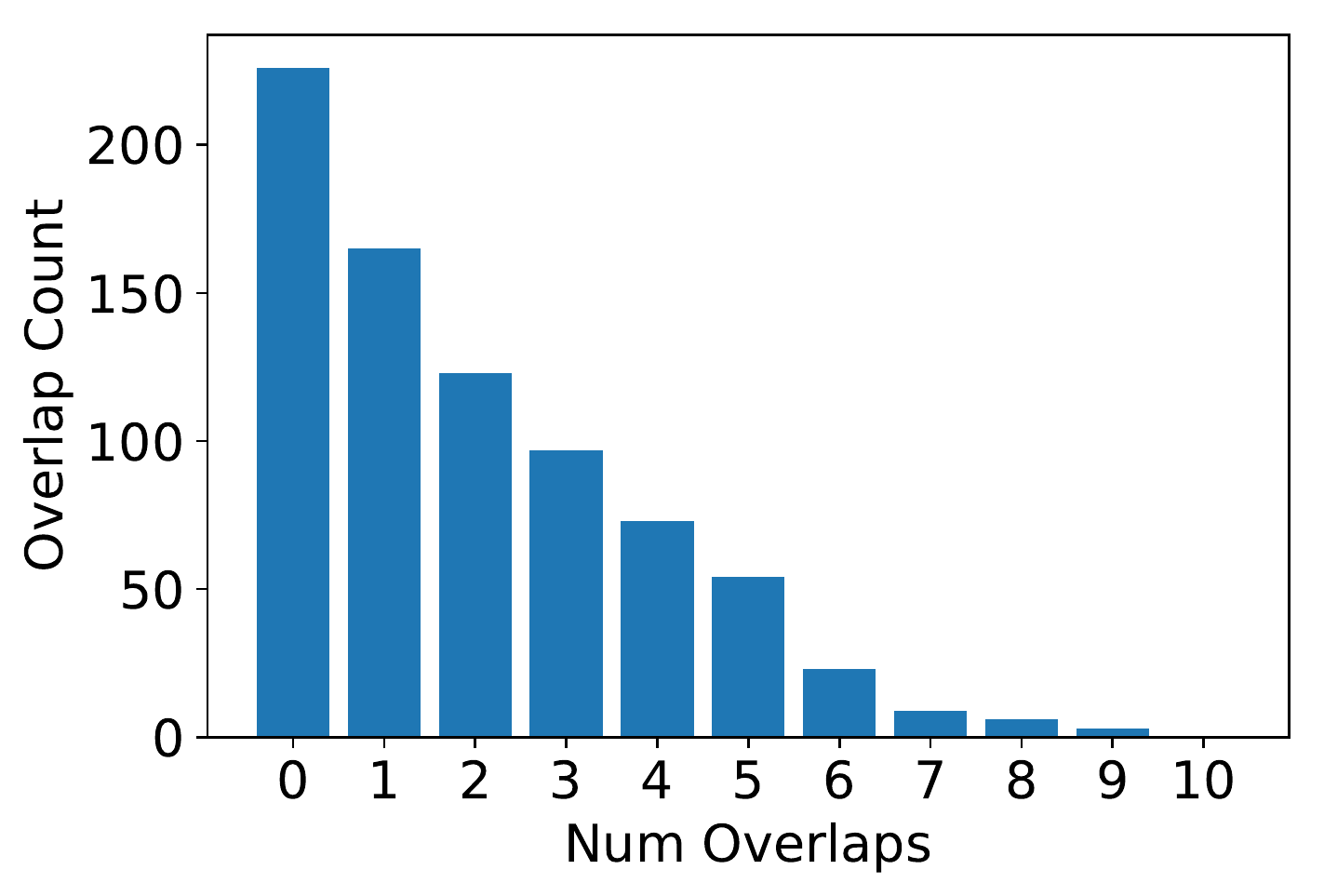}
  \caption{Performance of Doc2vec vector-space mapping relative to centralized learning, with Slack as private dataset, $sim_{k}=0.198$}
  \label{fig:vsm_doc2vec_centralized_slack}
\end{figure}

Figure~\ref{fig:vsm_doc2vec_centralized_slack} shows the distribution of the number of overlaps between the centralized case and the vector-space mapping case ($sim_{10} \times 10$). Here, the vector-space mapping algorithm is applied with randomly partitioned Slack data on each edge site as the private dataset on which a local Doc2vec model is trained and the 20-newsgroup dataset used to train the MLP mapper model. As seen in the distribution as well as indicated by the $sim_{k}$ of $0.198$, on average, the vector-space mapping model found about $2$ of the same users returned by the centralized model. Although this is inferior to the joint-learning case, it is an encouraging result showing that the mapping is meaningful, especially given the challenging setting of independently trained Doc2vec models.

However, there is a deeper insight here. Because the slack dataset as a private corpus was much smaller relative to the public corpus used, our hypothesis is that the Slack dataset did not have a rich enough vocabulary to map to the rich vocabulary of 20-newsgroup. The difference in corpus size is substantial, Slack having more than 1500 documents (one per user) whereas 20-newsgroup having more than 18000. To test this hypothesis, we switch to using Wikipedia as the private dataset, which is much richer in vocabulary.

\subsubsection{Wikipedia as private data}
\begin{figure}[htb!]
  \includegraphics[width=\columnwidth]{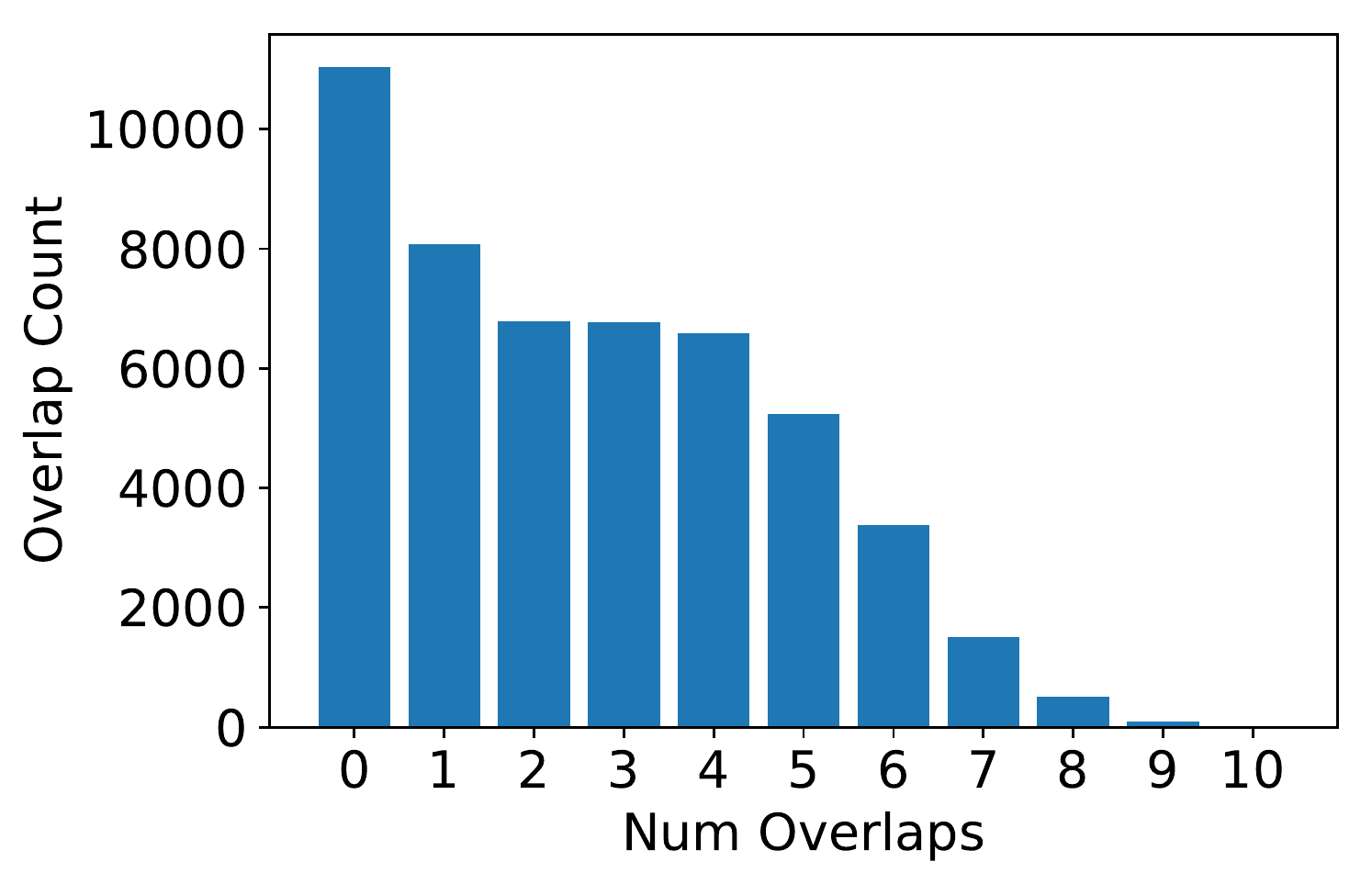}
  \caption{Performance of Doc2vec vector-space mapping relative to centralized learning with Wikipedia dataset, $sim_{k}=0.261$}
  \label{fig:vsm_doc2vec_centralized}
\end{figure}

Figure~\ref{fig:vsm_doc2vec_centralized} shows the distribution of the number of overlaps between the centralized case and the vector-space mapping case ($sim_{10} \times 10$). Here, the vector-space mapping algorithm is applied with randomly partitioned Wikipedia data on each edge site as the private dataset on which a local Doc2vec model is trained and the 20-newsgroup dataset used to train the MLP mapper model. As seen in the distribution as well as indicated by the $sim_{k}$ of $0.261$, on average, the vector-space mapping model found more than $2$ of the same users returned by the centralized model, which is better than the result with Slack data. Hence, we conclude that having rich vocabulary in the private dataset helps the mapper model to learn a more accurate mapping.

%\begin{itemize}
%    \item Federated doc2vec on slack data (overlap with centralized case)
%    \item Federated doc2vec on slack data (overlap with survey)
%    \item vector-space mapping doc2vec (private: wikipedia, public: newsgroup) (illustrative result: centralized, anarchisam, nintendo)
%    \item vector-space mapping doc2vec (private: wikipedia, public: newsgroup) (overlap with centralized case)
%    \item Federated node2vec on slack data graph, no node copy (overlap with centralized case)
%    \item Federated node2vec on slack data graph, with node copy (overlap with centralized case)
%    \item Federated node2vec on slack data, no node copy (overlap with survey)
%    \item Federated node2vec on slack data, with node copy (overlap with survey)
%\end{itemize}

\section{Related Work}\label{sec:related}
Main contributions of this paper borrowed and adapted key ideas from previous work in the areas of federated learning, semantic vector embedding, and edge resource representation and discovery. However, since this paper introduced new problems as well as novel algorithms to address them via a combination of the techniques from above areas, directly relevant prior work is limited. In the following, we describe related works in the broader areas of federated learning, semantic vector embedding, and edge resource management.

\subsection{Federated learning}
Federated learning is an actively evolving area of research that is receiving great attention in the edge computing community as it directly addresses the challenge of tapping into edge data in a privacy-preserving fashion \cite{kairouz-federated-2019}. One area of relevance is that of collaborative personalization where private user embedding is kept on device and combined with federated learning to personalize the predictions of a given model to the user \cite{bui-furl-2019}. Another idea in achieving greater personalization of a model is an empirical approach wherein each user device fine tunes a baseline model with local data and reports the differences in performance of the baseline model and the fine tuned model to allow the baseline model to be fine tuned with very different use cases of keyboard next-word recommendation and news recommendation \cite{wang-personalization-2019,qi-fedrec-2020}. 

Another body of work focuses on the performance optimization of the federated learning process to cater to resource-constrained applications. One idea is to minimize the size of the weight-updates reported by participating sites by either reporting only a subset or by computing a semantic sketch, to save communication costs \cite{mcmahan-comm-2016}. Another idea is an online learning approach to trade-off the performance with a constrained time-budget via allowing multiple local weight updates before a report is made to the server as well as achieving sparse gradient updates \cite{Wang-time-budget-2019,Wang-sparsification-2020}.

Whereas these works offer privacy-preserving and efficient techniques for learning joint models of user preferences, they assume that joint-learning is always possible. Even when joint learning is possible, they do not address the representation learning problem unlike this paper.

\subsection{Semantic vector embedding}
Semantic vector embedding techniques have been around for several years and are actively being developed primarily in the area of language modeling, e.g., BERT \cite{BERT-devlin-2018}, due to the availability of vast amounts of natural language data combined with their ability to learn semantic embedding in an unsupervised fashion. For instance, BERT-based gender stereotype identification was proven to outperform well-established lexicon-based approaches \cite{Zhao-gender-2020}. Similar to our motivating example of semantic search for people, vector embedding outperforms statistical approaches such as topic modeling in vector embedding of expertise \cite{han-distributed-2016}. Further, great progress has been made in learning vector embedding for a wide variety of domains. One intriguing example is embedding of neural-network architectures themselves to facilitate a search over the most suitable model architectures \cite{Zhang-arch-embedding-2020}. Another application is in semantic embedding of videos to enable efficient indexing and search over videos across edge and Cloud \cite{nahrstedt-sieve-2020}. 

The above advances in the vector embedding algorithms are promising and are complementary to the contributions of this paper. However, none of these efforts address the challenges of vector embedding on data locked into edges.

\subsection{Edge resource management}
Although resource discovery and management is not the focus of this paper, it helps to show the recent trend in this area towards applying leveraging machine learning as well as vector embedding techniques. Traditionally, edge resource management has been addressed via collection and maintenance of resource metadata \cite{salem-kinaara-2017,murturi-edge-2019}, in combination of with forecasting of resource demands and supply and optimizing resource allocation along a variety of objectives \cite{boubin-resources-2019}. A significant body of work also investigates economic paradigms in achieving optimal resource allocations among autonomous edges \cite{gua-mec-resource-2019,sun-double-2018}. More recently however, there has been an increasing recognition and validation of the black-box learning methods in managing resource discovery and allocation \cite{shi-resource-nn-2020,Wang-deep-resource-2019}. Such a trend clearly points to the promise of the research direction advocated in this paper.

\section{Conclusions and Future Work}\label{sec:conclusion}
With the increasing regulation and the growth in data originating at the edge, edge computing is poised to be a critical area of research with significant impact on how the IT systems are developed, deployed, and managed. This paper introduced the novel research direction of federated semantic vector embedding, building on the unique combination of the well-known techniques of federated learning and semantic vector embedding. Specifically, two research problems were formulated to cater to two separate settings in which edge sites want to collaborate in performing global semantic search across sites without sharing any raw data. 

The first setting, called joint-learning, is when the edge sites have a tightly coupled collaboration to participate in a synchronous joint-learning process and have an agreement on the model architecture, training algorithm, vector dimensionality, and data format. A novel algorithm to address the joint-learning problem is presented with the novel idea of vocabulary aggregation before starting the iterative federated learning process.

The second setting, called vector-space mapping, is when the edge sites do not agree on the various parameters of joint-learning or cannot participate in a synchronous process as they may need to join and leave dynamically. This is clearly a challenging setting and one of great significance in practice. Based on the novel idea of training another model to learn the mapping between vector-spaces based on a public dataset from any domain, an algorithm for addressing the vector-space mapping problem was presented.

Experimental evaluation using multiple natural language as well as graph datasets show that these algorithms show promising results for both algorithms compared to the baseline centralized case where all data can be aggregated on one site. Several important research questions remain open. How do these algorithms scale in the number of edge sites, differences in data distributions and amount of data at edge edge site? How do we interpret such semantic vectors and explain the similarity results they produce? This is the first paper in the area of federated semantic vector embedding and has unlocked several important research challenges for future research.

%\noindent\textbf{Acknowledgements}
%This research was sponsored by the U.S. Army Research Laboratory and the U.K. Ministry of %Defence under Agreement Number W911NF-16-3-0001. The views and conclusions contained in this %document are those of the authors and should not be interpreted as representing the official %policies, either expressed or implied, of the U.S. Army Research Laboratory, the U.S. %Government, the U.K. Ministry of Defence or the U.K. Government. The U.S. and U.K. Governments %are authorized to reproduce and distribute reprints for Government purposes notwithstanding %any copyright notation hereon.

\bibliographystyle{IEEEtranS}
\bibliography{sec_2020}

\end{document}